\documentclass[sigconf,authorversion=true]{acmart}

\usepackage{booktabs} % For formal tables

\usepackage[T1]{fontenc}
\usepackage{lmodern}
\usepackage{fixltx2e}
\usepackage{amsmath}
\usepackage{amssymb}
\usepackage{makecell}
\usepackage{caption}
\usepackage{wrapfig,lipsum,booktabs}
\usepackage{subfig}
\usepackage{graphicx}
\usepackage{url}
\usepackage[algoruled,boxed,linesnumbered]{algorithm2e}
\usepackage{xcolor} % http://www.ctan.org/tex-archive/macros/latex/contrib/xcolor

\usepackage{mathtools}
\usepackage{verbatim}
\usepackage{multirow}
\usepackage{paralist}
\usepackage{listings}
\usepackage{tikz}
\usepackage{dirtree}
\usepackage{tabularx}
\usepackage{tikz}
\usepackage[framemethod=default]{mdframed}
\usepackage{showexpl}
\usepackage{todonotes}
\graphicspath{{figures/}}
\newcommand{\nodes}{V}%{S_{PM}^a}
\newcommand{\arcs}{E}%{S_{PM}^r}
\newcommand{\freq}{\gamma}

\def\HiLiY{\leavevmode\rlap{\hbox to \hsize{\color{green!18}\leaders\hrule height .8\baselineskip depth .5ex\hfill}}}
\def\HiLiR{\leavevmode\rlap{\hbox to \hsize{\color{red!18}\leaders\hrule height .8\baselineskip depth .5ex\hfill}}}
\def\HiLiB{\leavevmode\rlap{\hbox to \hsize{\color{blue!18}\leaders\hrule height .8\baselineskip depth .5ex\hfill}}}

\copyrightyear{2018}
\acmYear{2018}
\setcopyright{acmcopyright}
\acmConference[ICSSP '18]{International Conference on the Software and
Systems Process 2018 (ICSSP '18)}{May 26--27, 2018}{Gothenburg, Sweden}
\acmBooktitle{ICSSP '18: International Conference on the Software and
Systems Process 2018 (ICSSP '18), May 26--27, 2018, Gothenburg, Sweden}
\acmPrice{15.00}
\acmDOI{10.1145/3202710.3203154}
\acmISBN{978-1-4503-6459-1/18/05}

\begin{document}
\title{Discovering Process Maps from Event Streams}
%\titlenote{Produces the permission block, and
 % copyright information}
%\subtitle{Extended Abstract}
%\subtitlenote{The full version of the author's guide is available as
%  \texttt{acmart.pdf} document}

\author{Volodymyr Leno}
\affiliation{\institution{University of\\ Tartu, Estonia}}
\email{leno@ut.ee}

\author{Abel Armas-Cervantes}
\affiliation{\institution{University of\\ Melbourne, Australia}}
\email{abel.armas@unimelb.edu.au}

\author{Marlon Dumas}
\affiliation{\institution{University of\\ Tartu, Estonia}}
\email{marlon.dumas@ut.ee}

\author{Marcello La Rosa}
\affiliation{\institution{University of\\ Melbourne, Australia}}
\email{marcello.larosa@unimelb.edu.au}

\author{Fabrizio M. Maggi}
\affiliation{\institution{University of\\ Tartu, Estonia}}
\email{f.m.maggi@ut.ee}

\renewcommand{\shortauthors}{V. Leno et al.}

\begin{abstract}
%Process mining is a body of methods to analyze event logs produced during the execution of business processes in order to extract insights for their improvement.
Automated process discovery is a class of process mining methods that allow analysts to extract business process models from event logs. Traditional process discovery methods extract process models from a snapshot of an event log stored in its entirety. In some scenarios, however, events keep coming with a high arrival rate to the extent that it is impractical to store the entire event log and to continuously re-discover a process model from scratch. Such scenarios require online process discovery approaches. Given an event stream produced by the execution of a business process, the goal of an online process discovery method is to maintain a continuously updated model of the process with a bounded amount of memory while at the same time achieving similar accuracy as offline methods. However, existing online discovery approaches require relatively large amounts of memory to achieve levels of accuracy comparable to that of offline methods. Therefore, this paper proposes an approach that addresses this limitation by mapping the problem of online process discovery to that of cache memory management, and applying well-known cache replacement policies to the problem of online process discovery. The approach has been implemented in .NET, experimentally integrated with the Minit process mining tool and comparatively evaluated against an existing baseline using real-life datasets.
\end{abstract}

%\begin{CCSXML}
%<ccs2012>
 %<concept>
 % <concept_id>10010520.10010553.10010562</concept_id>
 % <concept_desc>Computer systems organization~Embedded systems</concept_desc>
 %<concept>
%  <concept_id>10010520.10010575.10010755</concept_id>
%  <concept_desc>Computer systems organization~Redundancy</concept_desc>
%  <concept_significance>300</concept_significance>
%  <concept_id>10010520.10010553.10010554</concept_id>
%  <concept_desc>Computer systems organization~Robotics</concept_desc>
%  <concept_significance>100</concept_significance>
%  <concept_id>10003033.10003083.10003095</concept_id>
%  <concept_desc>Networks~Network reliability</concept_desc>
%  <concept_significance>100</concept_significance>
%</ccs2012>
%\end{CCSXML}

%\ccsdesc[500]{Computer systems organization~Embedded systems}


\maketitle

% !TEX root = ../paper.tex
\section{Introduction}\label{sec:introduction} %Marlon

Contemporary enterprise systems collect and maintain detailed data about the execution of the business processes they support. In particular, it is common to find records of business process events in Customer Relationship Management (CRM) systems, Enterprise Resource Planning (ERP) systems and other packaged enterprise systems.

Process mining~\cite{process-mining-book-2011} is a family of methods that allow users to exploit such records of business process events in order to gain insights into the performance of business processes and their conformance with respect to normative requirements.
Among other things, process mining methods allow users to automatically construct a process model from a given collection of event records (i.e., an \emph{event log}) generated by the execution of a business process.

%Problem area: process mining, automated process discovery -
The bulk of existing automated process discovery methods are designed to generate a process model from a snapshot of an event log stored in its entirety. In some scenarios, however, events are generated at a high throughput, to the extent that it is impractical to store the entire event log and to continuously re-discover a process model from scratch every time that new events arrive.
The latter scenarios require \emph{online}\footnote{The techniques that support online analysis of process data are also known as operational decision support tools \cite{DBLP:conf/apn/WestergaardM11,DBLP:conf/fase/MaggiMA12,DBLP:journals/eis/MaggiW17}.} process discovery approaches. Given an event stream produced by the execution of a business process, the goal of an online process discovery method is to maintain a continuously updated model of the process with low memory requirements and fast update times. One of the challenges of online process discovery is that of striking a tradeoff between the amount of memory and CPU consumption on the one hand, and the accuracy of the discovered process models on the other. Indeed, if we accept lower accuracy, we can keep in memory a selected subset of the observed behavior, and thus achieve lower resource consumption. On the other hand, if we wish to have an accurate process model, we need to somehow store all the behavior that has been observed throughout the event stream.

%most automated process discovery techniques and tools focused on discovering a process model from a full event log. In certain application areas however the process model needs to be continuously updated as new events arrive

%the same time achieving similar accuracy as offline methods.

%Problem statement: to discover continuously updated process models from event streams with low memory requirements and in a high throughput setting...

%Every event in the stream represents an execution of an activity in the context of an instance of the business process under observation.
This paper addresses the problem of online discovery of a specific type of process models called \emph{process maps}\footnote{Also known as directly follows graphs or DFGs~\cite{WeijtersAM06}.}. A process map is a directed graph where each observed activity is represented as a node, and an arc between activities A and B denotes the fact that activity B has been observed immediately after activity A in at least one instance of the process (a.k.a.\ \emph{case}). An arc between A and B is annotated by the number of times that B \emph{directly follows} A. Process maps are a popular representation in the context of process mining. They are supported by virtually all commercial tools in the field, including Celonis\footnote{\url{http://www.celonis.com}}, Disco\footnote{\url{https://fluxicon.com/disco/}}, Minit\footnote{\url{http://http://minitlabs.com/}} and MyInvenio\footnote{\url{http://www.my-invenio.com/}}. They are also supported by the so-called Fuzzy miner plug-in of ProM, an open-source process mining toolset\footnote{\url{http://www.promtools.org/}}, and they are the starting point for several other automated process discovery techniques such as the Heuristics Miner and the Inductive Miner~\cite{process-mining-book-2011}.

While the problem of discovering a process map is well understood in the setting where the entire log is available at once, it has been less studied in the case where events arrive one by one, and where we need to produce an up-to-date version of the process map after every event arrival.
%In this context, re-computing the process map from all previously observed events is computationally inefficient. Moreover, in some scenarios, the number of events in the stream is so high, that it is not practical to store in memory all the events that have been observed since the beginning of the stream.
In this latter setting, the challenge is to incrementally update the process map when a new event arrives, while minimizing the amount of memory. A previous approach to this problem, namely Lossy Counting with Budget (LCB)~\cite{BurattinSA14}, constructs and maintains a process map incrementally with a fixed amount of memory (the so-called \emph{memory budget}). However, unless the budget is set very high (and it is unclear how high), this approach leads to loss of accuracy, meaning that the process map produced at a given point in time is not identical to the one that would be calculated if the entire set of events was available at once. In other words, in the LCB approach, there is no clear way to control the accuracy of the resulting process map.

Therefore, in this paper, we propose an alternative approach inspired by cache replacement policies used in the field of cache management (e.g., Least-Recently-Used - LRU and Least-Frequently-Used - LFU). The idea is to keep in memory the last event of every ``ongoing'' trace, and a subset of the process map up to a certain memory budget, meaning that if the size of the process map exceeds the available memory, some arcs and nodes in the process map may be ``deleted'' to create more space. The resulting approach hence requires a fixed memory budget for the process map plus a variable amount of memory to maintain the last event of each ongoing case (the latter amount of memory is bounded by the maximum number of cases that can be ``active'' at any given point in time). The paper also provides a formula that tells us what memory budget is required if the goal is to construct the process map in a lossless manner. Three variants of the proposed approach are studied in the paper, each one corresponding to a different cache replacement policy.

An experimental evaluation using real-life event logs is reported. This evaluation shows that the approaches based on cache replacement policies outperform the existing LCB approach when the objective is to obtain a lossless or near-lossless process map. Specifically, these approaches can incrementally maintain a process map with an accuracy of 90\% or above with significantly less memory usage than what the LCB approach requires to achieve the same accuracy.

The rest of the paper is structured as follows. Section 2 provides a summary of related work in the field of online process discovery and an overview of the cache replacement policies used in our approach. Section 3 presents the proposed approach, while Section 4 discusses its empirical evaluation. Finally, Section 5 summarizes the contributions of the paper and outlines directions for future work.

% !TEX root = ../paper.tex
\section{Background and related work}\label{sec:background} %Fabrizio

This section briefly presents the problem of online process discovery, reviews existing related work, and introduces relevant concepts on cache replacement policies.

\subsection{Online process discovery}
Automated process discovery is an umbrella term used to refer to techniques that generate structured process descriptions (process models) from a set of business process event records.
Starting from \cite{Agrawal1998:Mining}, a plethora of techniques have been proposed in this field.
Detailed surveys and empirical evaluations of existing techniques in this field are reported in~\cite{WeerdtBVB12,Augusto2017Survey}.
The vast majority of the techniques proposed in the field assume that the input is an event log consisting of a collection of event records available all at once (offline). Only a handful of previous studies have addressed the problem of online discovery of process models from streams of events, which is the scope of this paper. 
%They can be classified as pure algorithmic, heuristic and genetic \cite{process-mining-book-2011}.

%More recently, several works have been published
%starting from the awareness that techniques based
%on declarative languages are suitable for the discovery of unpredictable, variable processes working in unstable
%environments \cite{Chesani2009:Mining, CiccioM12, MaggiCIDM, Maggi2012,Maggi2013}.

%Process stream mining consists in the extraction of process structures from continuous and rapid process data records. Even if, in the last years, dozens of process discovery techniques have been proposed \cite{process-mining-book-2011}, these techniques all work on static event logs and not on streaming data. Only few works in process mining aim at mining event streams.

One of the early studies addressing the problem of online discovery of process models is that of Kindler et al.~\cite{Kindler2005a,Kindler2006c}, which addresses the following problem: given a process model (specifically a Petri net) representing the observed behavior up to a certain point in time, and given a set of process execution traces observed during a time window, compute a new version of the process model taking into account the newly observed events. In other words, the problem addressed is that of incremental process model refinement. However, the authors do not address this problem in an event streaming setting, since they do not take into account memory limitations. Specifically, the approach in~\cite{Kindler2005a,Kindler2006c} maintains a process model capturing all the behavior observed in the event stream, no matter how large this model becomes. Also, in their approach, the model is not updated after every event, but only when a new case has completed (the approach takes completed traces as input). Finally, their approach relies on computationally demanding model merging techniques that are not designed for high-throughput settings. Similar remarks can be made about a related study~\cite{Sharp2008}, which proposes an approach for incremental translation of transition systems into Petri nets.

Closer to our work is that of Burattin et al.~\cite{Burattin2012b,BurattinSA14}, which addresses the problem of online discovery of process maps under limited memory. The authors propose three approaches to solve this problem. The first one is a sliding window-based approach, wherein only the last N events (window) in the stream are maintained, and the process map is computed based on this window. The authors show that this simple approach is not sufficiently efficient in a streaming setting. The second approach is an adaptation of Lossy Counting (LC)~\cite{MankuM02}, a general approach for maintaining item counts over event streams. The item counts are based on items stored in partitions called ``buckets''. The idea of LC is to count how many times the items (e.g., activities or directly follows relations between pairs of activities) have been observed in a bucket of a given size. When the maximum size of the bucket is reached, infrequent items are cleaned up. The accuracy of LC can be controlled by a user-defined error margin $\epsilon \in [0..1]$. The authors adapt the LC approach in order to take into account the fact that the activity instances (events) in a business process event stream refer to multiple cases. Hence, each time an event arrives with reference to a given case $c$, the directly follows relation is updated by looking at the last event that occurred in case $c$. This requires the storage of the last observed event of each case.
A disadvantage of LC is that there is no limitation on the memory employed. Hence, the authors in~\cite{Burattin2012b,BurattinSA14} outline a third approach called Lossy Counting with Budget (LCB) originally proposed in~\cite{DaSanMartino2012}. The idea of LCB is to use buckets of variable size (as opposed to fixed size as in LC). A bucket is considered full only when the maximum available memory (the budget) is reached, then the same cleanup procedure as in LC is applied. However, if the cleanup procedure does not free-up space because no item fulfills some required conditions, then the cleanup conditions are relaxed until some items are deleted. Differently from LC, LCB cannot guarantee a given level of accuracy, however, LCB guarantees that the allocated memory budget is never exceeded.

The experimental evaluation reported in~\cite{Burattin2012b,BurattinSA14} shows that LCB has significantly better processing times (and naturally lower memory consumption) than LC. An improved version, solely with respect to processing times, of this approach has been proposed by Hassani et al.~\cite{hassaniSRS2015}. In \cite{DBLP:conf/otm/MaggiBCS13,BurattinCM14,DBLP:journals/tsc/BurattinCMS15}, LCB has been applied for the discovery of declarative process models expressed in terms of Declare constraints \cite{Pesic2007}. Based on the same approach, in \cite{vanZelstDA2017}, the authors develop a generic architecture that allows for adapting different classes of offline process discovery techniques to the online setting.
However, none of these mentioned approaches address the main limitation of LCB, namely that it is unclear how the memory budget should be set in order to achieve a given level of accuracy.
Our work aims at tackling this limitation by putting forward an alternative approach where the tradeoff between memory and accuracy can be clearly specified.

Burattin et al.~\cite{Burattin2012b,BurattinSA14} and van Zelst et al.~\cite{vanZelstDA2017} have noted that a process map obtained from an event stream can be used to generate a process model in other process modeling notations -- e.g., in the BPMN notation\footnote{\url{http://bpmn.org}} -- by periodically invoking a separate algorithm that discovers a process model from the latest version of the process map. This latter idea is complementary and orthogonal to the core idea of incrementally computing a process map. Indeed, if we know how to maintain a process map from an event stream, we can then periodically invoke any (incremental) algorithm that computes a BPMN-like process model from a process map. %Accordingly, in this paper we focus on the problem of online discovery of process maps.

%\todo{Add comparison with \cite{vanZelstDA2017}.}

%The first one is already an improvement over Burattin’s approach. The second is the work of Bas Zelst and is a generalization of Burattin’s approach. Note that the first paper improves over Burattin’s approach in terms of execution time, while the second paper improves in terms of number of internal states produced. None of them compares with Burattin’s in terms of memory consumption.
% Marwan Hassani, Sergio Siccha, Florian Richter, Thomas Seidl: Efficient Process Discovery From Event Streams Using Sequential Pattern Mining. SSCI 2015: 1366-1373
%https://link.springer.com/article/10.1007%2Fs10115-017-1060-2

%Another approach in the field of online process discovery is that of~ \cite{Maggi2013,BurattinCM14}, which outlines an approach for online discovery of declarative process models. \todo[inline, color=red]{Explain here what is fundamentally different between \cite{Maggi2013,BurattinCM14} and the problem addressed here. For sure, we can mention that the above techniques produce a different type of model (declarative versus procedural), but are there other fundamental conceptual differences in terms of the problem addressed, e.g., is the technique in \cite{Maggi2013,BurattinCM14} also applicable under limited memory and where a new version of the model is produced after every event arrival?}
%based on algorithms for data stream mining

%\todo{emphasize that the latter addresses the problem of discovering a different class of process models.}

The problem of online discovery of process models is also related to that of business process drift detection~\cite{jcthesis,DBLP:conf/bpm/MaaradjiDRO15,DBLP:conf/er/OstovarMRHD16,DBLP:conf/caise/OstovarMRH17}, which can be formulated as follows: given an event log covering a given time window, to detect points in this time window where the behavior of the underlying process has changed (so-called \emph{change points}), and identify specific changes that have occurred at those change points. However, note that drift detection techniques do not deal with the problem of discovering a process model, nor are they intended to update a process model incrementally.

\subsection{Cache replacement policies}

The web is the most important source of information and communication in the
world. The majority of web objects are static, therefore caching them at HTTP
proxies can reduce network traffic and response time. However, since the size
of the cache is limited, some strategies are needed to identify the objects that
have to be stored in the cache and the objects that have to be thrown away for
clearing space for new ones.
These strategies are called \emph{cache replacement policies}.
The cache replacement policies can be classified into three categories:
recency-based, size-based and frequency-based policies, which take into
account the web object properties (recency, size and frequency) for selecting
the elements to keep in or delete from the cache.
%In some cases, policies of one type can use methods from another type in order to
%choose the objects for removal (this can happen when the main characteristics of objects are
%equal).

Recency-based strategies are derived from a property known as temporal locality, i.e.,
the measure of how likely an object is to be requested again when it has been
requested within a certain time span. Thus the elements to remove are those that have not
been requested for the longest period of time.
%These strategies exploit the recency information for their victim selection process.
%Recency-based strategies are typically straightforward to implement and use queues and linked lists.
%The recency-based strategy we adapt and use in this paper is the Least Recently Used (LRU) strategy,
%which is one of the most known and commonly used strategies. It is based on the idea that the block
%in the cache that has not been used for the longest period of time should be replaced first.
Size-based strategies aim at minimizing the memory consumption,
through the removal of the largest objects. Frequency-based strategies use a property known as spatial locality, i.e., the likelihood that an object will appear again based
on how often it has been seen before. Other extensions to the frequency-based strategies
use other information to complement the frequency, e.g., they use an aging factor to delete some old
elements over time.
%In this paper, we use the Least Frequently Used (LFU) strategy, which removes the object
%with the smallest frequency of occurrence, and the LFU with Dynamic Aging (LFU-DA)
%strategy based on the same idea but less complex and easier to manage than LFU because
%it does not have parameters to be set.

In process discovery from event streams, the main challenge is that it is not possible to
keep the whole event stream in memory. Therefore, we adapt cache replacement policies
as a mechanism to ``free-up'' space once a certain limit is reached. Specifically, we apply
cache replacement policies for clearing the memory for upcoming data from a stream of events.
%We use the Least Frequently Used (LFU) strategy, which removes the object
Particularly, in this paper (Section~\ref{sec:approach}), we use recency-based and
frequency-based cache replacement policies for the discovery of process maps. 

\begin{figure}[t!]
\centering
\begin{minipage}[t][][b]{0.39\textwidth}
    \centering
    \scalebox{1.0}{
\begin{tabular}{|l|l|l|l|}
\hline
	{\bf Case id} & {\bf Activity name} & {\bf Timestamp} \\ \hline
	1 & Create Fine &19/04/2017  14:00:00 \\ \hline
	2 & Create Fine & 19/04/2017  15:00:00 \\ \hline
	1 & Send Bill & 19/04/2017  15:05:00 \\ \hline
	2 & Send Bill & 19/04/2017  15:07:00 \\ \hline
	3 & Create Fine & 20/04/2017  10:00:00 \\ \hline
	3 & Send Bill & 20/04/2017  14:00:00 \\ \hline
	4 & Create Fine & 21/04/2017  11:00:00 \\ \hline
	4 & Send Bill & 21/04/2017  11:10:00 \\ \hline
	1 & Process Payment & 24/04/2017  14:30:00 \\ \hline
	1 & Close Case & 24/04/2017  14:32:00 \\ \hline
	2 &  Send Reminder & 19/04/2017  10:00:00 \\ \hline
	3 & Send Reminder & 20/05/2017  10:00:00 \\ \hline
	2 & Process Payment & 22/05/2017  9:05:00 \\ \hline
	2 & Close Case & 22/05/2017  9:06:00 \\ \hline
	4 & Send Reminder & 21/05/2017  15:10:00 \\ \hline
	4 & Send Reminder & 21/05/2017  17:10:00 \\ \hline
	4 & Process Payment & 26/05/2017  14:30:00 \\ \hline
	4 & Close Case & 26/05/2017  14:31:00 \\ \hline
	3 & Send Reminder & 20/06/2017  10:00:00 \\ \hline
	3 & Send Reminder & 20/07/2017  10:00:00 \\ \hline
	3 & Process Payment & 25/07/2017  14:00:00 \\ \hline
	3 & Close Case & 25/07/2017  14:01:00 \\ \hline
\end{tabular}
} \label{fig:log}
    \caption*{Event log}
    \end{minipage}
%\hspace{2mm}
  \begin{minipage}[t][][b]{0.49\textwidth}\centering
 \vspace{5mm}
\includegraphics[scale=0.4]{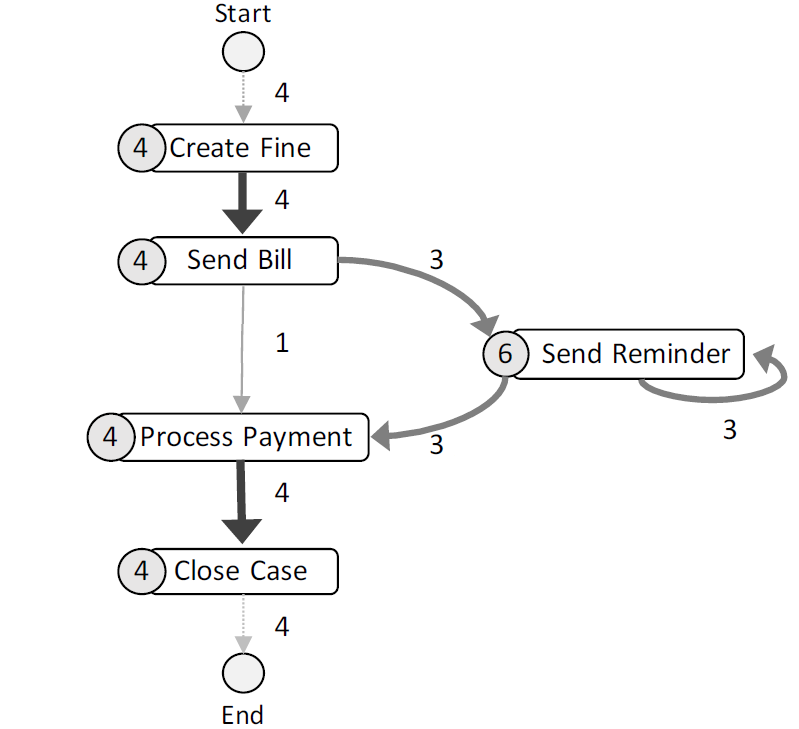}
%\vspace{6mm}
\caption*{Directly follows graph}
\label{fig:proc:map}
\end{minipage}
\caption{Event log and its corresponding process map}
\vspace{-4mm}
\label{fig:log:proc:map}
\end{figure}

\begin{figure*}[t]
\centering
\includegraphics[scale=0.5]{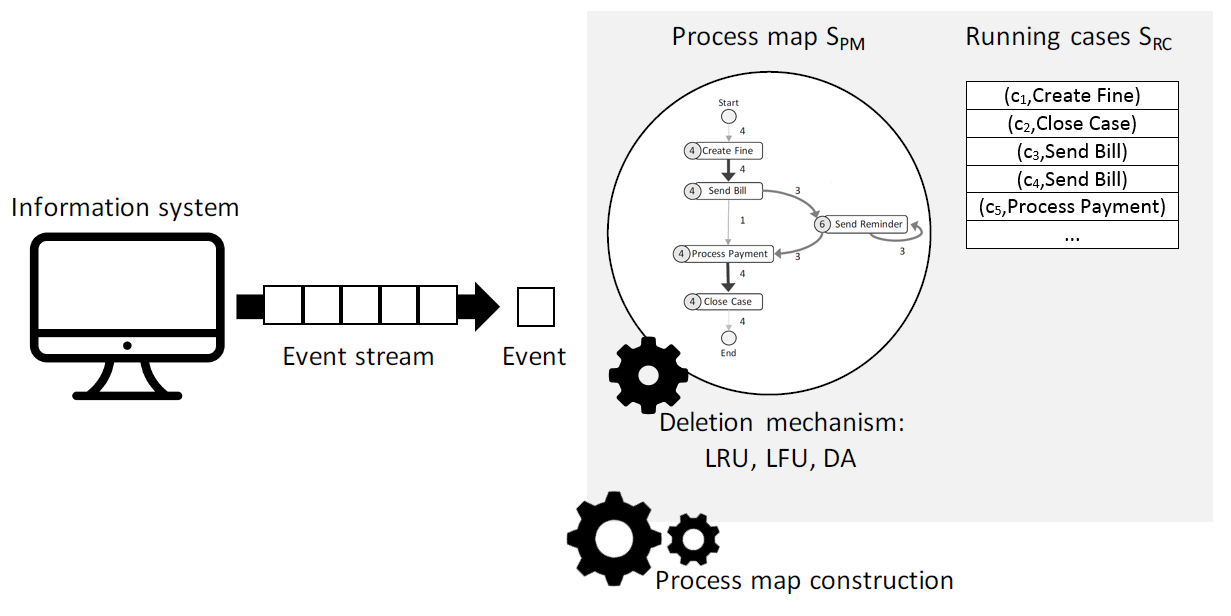}
\caption{Overview of the proposed approach}
\label{fig:overview}
%\vspace{-5mm}
\end{figure*}

% !TEX root = ../paper.tex
\section{Approach}\label{sec:approach} %Abel

This section presents the main contribution of the paper. The first subsection introduces the fundamental notions of event logs and process maps, while the second subsection presents a novel technique for the discovery of process maps from event streams.

\subsection{Preliminaries}

\emph{Event logs} record the execution of activities in a business process. Every execution of a process constitutes a \emph{case} and produces a sequence of activity occurrences called a \emph{trace}.  The activity occurrences are referred to as \emph{events} and can have several attributes. Here, we assume that events are equipped with only three attributes: case id, activity name and timestamp. For instance, Fig.~\ref{fig:log:proc:map} shows an event log (left-hand side) with 5 activities and 22 events.

\begin{definition}[Event]
Let $\mathcal{A}$ be the set of all activity names, $\mathcal{C}$ be the set of case identifiers and $\mathcal{T}$ be the set of timestamps of an event log. An \emph{event} is a tuple $e = \langle A,C,T \rangle \in \mathcal{A}  \times \mathcal{C} \times \mathcal{T}$ and represents the occurrence of activity $A$ at time $T$ in the case with id $C$. The set of all possible events is denoted as $\mathcal{E}$.
\end{definition}

Given an event $e = \langle A, C, T \rangle$, we refer to its activity, case and timestamp as $e|_{A} = A$, $e|_{C} = C$ and $e|_{T} = T$, respectively. The timestamps reflect the order of execution of the events. The bounded sequence of ordered events (w.r.t. timestamps) occurring within the same case is called a \emph{trace}; whereas, an unbounded sequence of events belonging to different cases is called an \emph{event stream}. %Figure~\ref{fig:log:proc:map} displays 4 different cases, and thus 4 traces, e.g., the trace with case id 1 is the sequence of events $\langle Create\ Fine, Send\ Bill, Process\ Payment, Close\ case \rangle$.

\begin{definition}[Trace and event stream]
Let $\mathcal{E}$ be the set of events of an event log. A \emph{trace} is a bounded sequence of events $\sigma = \langle e_1, \dots, e_n \rangle$, such that $e_x|_T < e_y|_T$ and $e_x|_C = e_y|_C$, for any $1 \leq x < y\leq n$ and $e_x, e_y \in \mathcal{E}$. An \emph{event stream} is an unbounded sequence of events $\sigma = \langle e_1, \dots \rangle$ with the same order among the events as in a trace, i.e., $e_x|_T < e_y|_T$, for any $1 \leq x < y$.
\end{definition}

Two events $e_x$, $e_y$ are said to be in a \emph{directly follows} relation if they belong to the same case and they occurred consecutively. %, i.e., $e_x|_C = e_y|_C$, $e_x|_T < e_y|_T$ and $\nexists e_z : e_x|_T < e_z|_T < e_y|_T$.

\begin{definition}[Directly follows relations]
Let $\sigma = \langle e_1, e_2,\allowbreak \dots \rangle$ be a sequence of events, either a trace or an event stream.
A pair of events $e_x$ and $e_y$ are in a \emph{directly follows} relation, denoted as $e_x \Rightarrow e_y$, if $e_x|_C = e_y|_C$, $e_x|_T < e_y|_T$ and $\nexists e_z \in \mathcal{E}: e_z|_C = e_x|_C = e_y|_C ~\land~ e_x|_T < e_z|_T < e_y|_T$, for any $1 \leq x, y, z$.
\end{definition}

Trace $\langle Create\ \allowbreak Fine,~\allowbreak Send\ \allowbreak Bill,~ \allowbreak Process\ \allowbreak Payment,\allowbreak Close\ \allowbreak Case \rangle$ contains three directly follows relations: $Create\ \allowbreak Fine\allowbreak \Rightarrow\allowbreak Send\ \allowbreak Bill$,
$Send\ \allowbreak Bill\allowbreak \Rightarrow\allowbreak Process\ \allowbreak Payment$ and $Process\ \allowbreak Payment\allowbreak \Rightarrow\allowbreak Close\ Case$.

The directly follows relations implicitly represented in a set of traces or an event stream can be abstracted into a \emph{directly follows graph}, also called a \emph{process map}. This abstraction is used by some process mining algorithms, such as the Heuristic Miner~\cite{WeijtersAM06}, as a baseline for the discovery of other types of models, e.g., models in BPMN notation. Intuitively, a directly follows graph is a graph where nodes represent activities, arcs represent directly follows relations, and every node and every arc is annotated with its frequency (number of times the activity/relation has been observed in the log).

\begin{definition}[Directly follows graph]
Let $\mathcal{A}$ be the set of all activity names of an event log. A directly follows graph is a tuple $S_{PM} =\langle \nodes , \arcs, \freq \rangle$, where $\nodes \subseteq \mathcal{A}$ is a set of nodes representing activities, $\arcs \subseteq \nodes \times \nodes$ is a set of arcs representing directly follows relations between activities, and $\freq: \nodes \cup \arcs \rightarrow \mathbb{N}_0$ is a function associating nodes and arcs to frequencies.
\end{definition}

Figure~\ref{fig:log:proc:map} displays an event log and its representation as a directly follows graph aside, nodes and arcs are annotated with their frequency, and a start and an end node were inserted to denote start and end of traces. In this graphical representation, the thickness of the arcs vary depending on their frequency, the thicker the arrow the higher the frequency.
%Note that, in this example, we assume that the entire event log is known, however in case of event streams, which is the problem tackled in this paper, the events can keep coming with a high arrival rate.

\subsection{Online process map discovery}

This subsection presents a novel technique for discovering process maps from event streams. The general idea is to maintain two memory partitions: $S_{PM}$, which is allocated to store the process map itself, and $S_{RC}$, which keeps track of the observed cases in the event stream. Then, for every event $e = \langle A, C, T \rangle$ observed in the event stream, our technique will seek for the last observed event $e'$ in $S_{RC}$ with the same case id $c$; if $e'$ exists then a new directly follows relation $e'|_{A} \Rightarrow e|_{A}$ is either created (if it does not exist already) or updated (if it exists) by increasing its frequency; finally $e$ is stored in $S_{RC}$. Note that every observed event triggers the update of the process map, either by creating a new node (if the activity has not been observed), or by creating a new arc, or by increasing the frequency of a node and/or an arc. An overview of the proposed technique is displayed in Fig.~\ref{fig:overview}

The storage of all the information read from an event stream can result impractical since new events, relations and cases can continuously emerge. In order to cope with possible memory limitations, $S_{RC}$ stores only the last observed event for every case; note that this information is sufficient to discover the directly follows relations. Furthermore, a memory size $B_{PM}$ and $B_{RC}$ -- hereinafter referred to as \emph{budgets} -- can be associated to each partition $S_{PM}$ and $S_{RC}$, respectively. Both budgets determine the amount of objects (activities, relations and cases) that can be allocated for the process map and for the running cases. Thus, in order to have a lossless representation of a process map from an event stream with activity names in $\mathcal{A}$, it is necessary to allocate enough space for storing a graph with $N$ nodes, where $N$ is the number of distinct activity names in $\mathcal{A}$, and all possible arcs between every pair of activities (including self-loops). Thus, for a lossless representation, the budget $B_{PM}$ has to be equal to:

\begin{equation}\label{eq:opbpmn}
	B_{PM} = \frac{N (N+1)}{2} + N.
\end{equation}

Intuitively, Equation~\ref{eq:opbpmn} counts the number of arcs in a clique, plus the self-loops and the number of nodes. However, if the budget $B_{PM}$ is not large enough to store a lossless process map representation, then once $S_{PM}$ is full, some elements need to be deleted to give place to new relations or activities observed in the stream.
The budget $B_{RC}$ controls the amount of running instances that can be stored in $S_{RC}$ and, similarly to $S_{PM}$, once it is full then some elements need to be deleted in order to memorize new observed cases. Both partitions can be maintained separately, have distinct sizes and have different deletion mechanisms.

We propose three deletion mechanisms for $S_{PM}$ based on well-known concepts from cache memory management. \emph{Cache replacement policies} identify elements to be cached or deleted in HTTP proxies to reduce network traffic and response time~\cite{ArlittCDFJ00}. %The cache replacement policies can be classified into three categories: recency-based, size-based and frequency-based policies.
%These policies can be used in combination to decide what elements to remove.
We adapt the three following cache replacement policies for the deletion mechanism for $S_{PM}$:
\begin{itemize}
%\scriptsize
	\item \emph{Least Recently Used (LRU)} deletes the elements in the cache that have not been used for the longest period of time. From the basics of temporal locality, the elements that have been referenced recently will be likely referenced in the near future. This policy works well when there is a high temporal locality of references in the workload. In our context, relations and activities read from the event stream are annotated with the ``last seen index'' attribute to keep track of their recency.

	\item \emph{Least Frequently Used (LFU)} removes the element with the smallest frequency of occurrence. In our context, no additional attributes are required since the frequency of the relations and activities is already stored as part of the directly follows graph through the use of $\freq$.

	\item \emph{LFU with Dynamic Aging (LFU-DA)} removes the element with the highest value given by $K_T = \freq(t) + L$, where $t$ is either a relation or an activity and $L$ is a dynamic aging factor. Initially, $L$ is set to 0, thus the policy is the same as LFU before the first deletion. However, once an element $t'$ is deleted, $L$ is set to $K_{t'}$.
\end{itemize}

%\begin{wrapfigure}{r}{0.45\textwidth}
\begin{figure}
%\vspace{-3mm}
\centering
\includegraphics[scale=0.55]{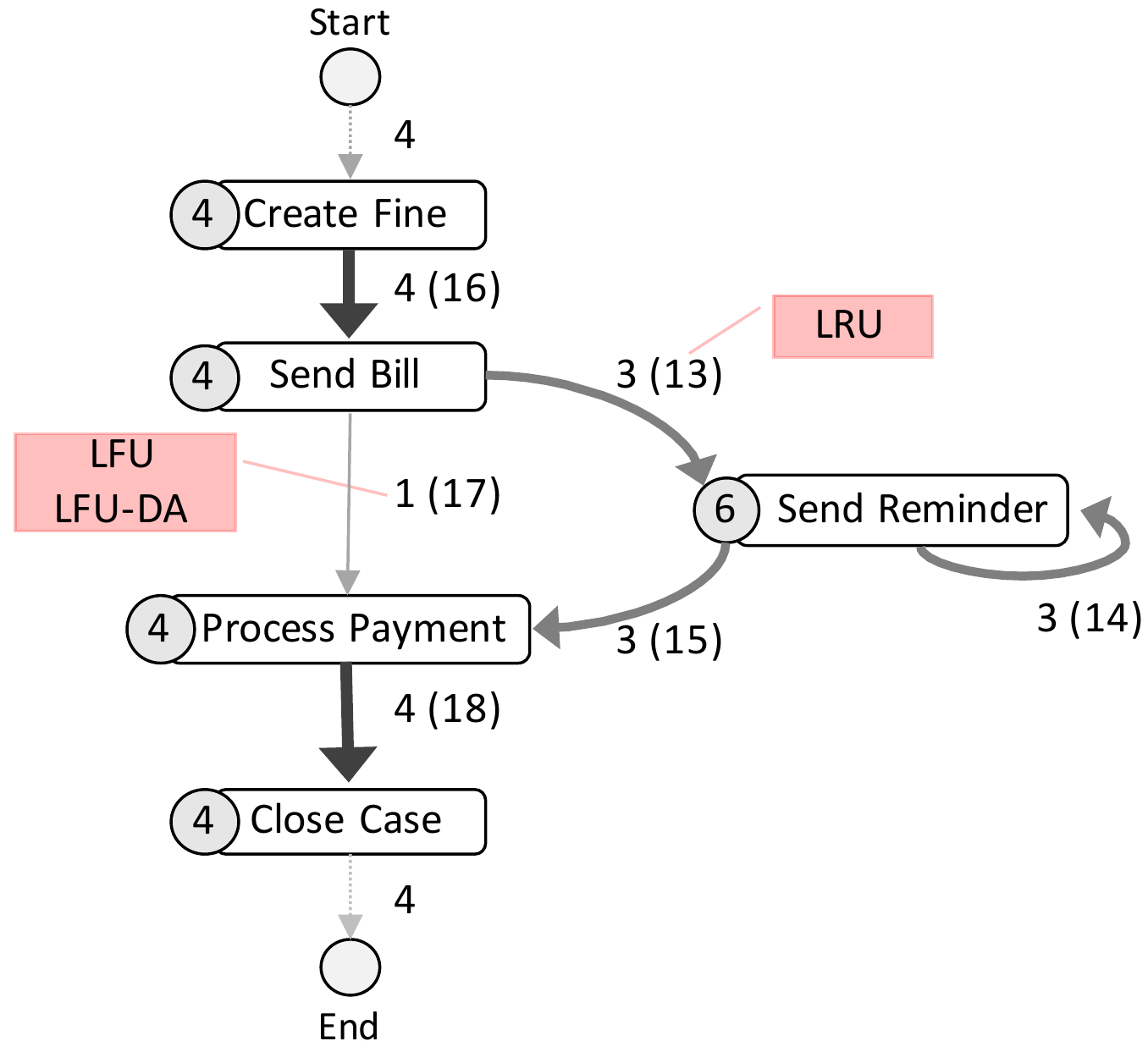}
\caption{Relations to be deleted}
\label{fig:del:strat}
%\vspace{-4mm}
\end{figure}
Once $S_{PM}$ is full and new information needs to be stored, the elements selected by the chosen policy are deleted from the partition $S_{PM}$. The deletion procedure considers two cases:
\begin{inparaenum}[(i)]
	\item if the element is a relation, then it is simply removed, but
	\item if it is an activity, then the activity and all the relations associated to it are removed.
\end{inparaenum}
$S_{PM}$ is subdivided into $\arcs$ and $\nodes$ for storing the relations and the activities, respectively. With abuse of notation, let $|\arcs|$ (resp.\ $|\nodes|$) denote the size of the partition for the relations (resp.\ for the activities), thus $B_{PM} = |\arcs| + |\nodes|$.

\begin{table*}[t!]
\centering
\begin{tabular}{lllll}
                                        &                       &  &  & \\
                                              &                       &  &  & \\
\multicolumn{1}{l|}{\multirow{3}{*}{\makeatletter
\newcommand{\removelatexerror}{\let\@latex@error\@gobble}
\makeatother

\begin{minipage}[t]{0.48\linewidth}
\vspace{-30mm}
\begingroup
\removelatexerror% Nullify \@latex@error
\begin{algorithm*}[H]{
  \scriptsize
  \SetKwInOut{Input}{input}
  \SetKwProg{Fn}{Function}{}{}
  \SetKwFor{Loop}{Loop}{}{EndLoop}
  \Input{Event stream $ES$, $B_{PM}$ and $B_{RC}$}
	$\arcs, \nodes, S_{RC} \leftarrow \emptyset$\\
	\HiLiR$idx \leftarrow 1$ \tcp{LRU}
	\HiLiB$L \leftarrow 1$ \tcp{LFU-DA}

\Loop{}{
	$e = \langle A, C , T \rangle \leftarrow observe(ES)\;$\\
	\uIf{$A \in \nodes$}{
		$\freq(A) = \freq(A) + 1$\\
		\HiLiR update $lastSeenIdx(A)$ \tcp{LRU}
	}\Else{
		\If{$B_{PM} = |\arcs| + |\nodes|$}{
			$deletionMechanism()$\\
			\HiLiB update $L$ \tcp{LFU-DA}
		}
		\HiLiR$\nodes \leftarrow \nodes \cup \{(A, idx)\}$ \tcp{LRU}
		\HiLiY$\nodes \leftarrow \nodes \cup \{A\}$ \tcp{LFU}
		\HiLiB$\nodes \leftarrow \nodes \cup \{(A, L)\}$ \tcp{LFU-DA}
	}
	
	\uIf{$C \in S_{RC}$}{
		$rc \leftarrow \langle C, A' \rangle\in S_{RC}$\\
		\uIf{$r = (A', A) \in \arcs$}{
			$\freq(r) = \freq(r) + 1$\\
			\HiLiR update $lastSeenIdx(r)$  \tcp{LRU}
		}\Else{
			\If{$B_{PM} = |\arcs| + |\nodes|$}{
				$deletionMechanism()$\\
				\HiLiB update $L$  \tcp{LFU-DA}
			}
			\HiLiR$\arcs \leftarrow \arcs \cup \{(r, idx)\}$ \tcp{LRU}
			\HiLiY$\arcs \leftarrow \arcs \cup \{r\}$ \tcp{LFU}
			\HiLiB$\arcs \leftarrow \arcs \cup \{(r,L)\}$ \tcp{LFU-DA}
		}
		$S_{RC} \leftarrow (S_{RC} \backslash \{  rc \}) \cup \{(c, A)\}$
		
		\If{$isExpired(c)$}{
			$S_{RC} \leftarrow S_{RC} \backslash \{(C, A'')\}$ for any $A''$
		}
	}\Else{
		$S_{RC} \leftarrow S_{RC} \cup \{(c, A)\}$
	}
	$idx \leftarrow idx + 1$
}
 \caption{Process map construction}\label{alg:process:map}
}
\end{algorithm*}
\endgroup
%}

\end{minipage}

}} &  & {% !TEX root = ../paper.tex

\makeatletter
\newcommand{\removelatexerror}{\let\@latex@error\@gobble}
\makeatother

\begin{minipage}{0.45\linewidth}
\begingroup
\removelatexerror% Nullify \@latex@error

\begin{algorithm*}[H]{
  \scriptsize
  \SetKwInOut{Input}{input}
  \SetKwProg{Fn}{Function}{}{}
  \SetKwFor{Loop}{Loop}{}{EndLoop}
  \Input{$\arcs, \nodes$}

	$min_a \leftarrow min\{ idx | (A, idx) \in \nodes \}$\\
	$min_r \leftarrow min\{ idx | (r, idx) \in \arcs \}$\\
	\uIf{$min_a > min_r$}{
		$\arcs \leftarrow \arcs \backslash \{ (r, idx) \}$
	}\Else{
		$\nodes \leftarrow \nodes \backslash \{ (A, idx) \}$\\
		\For{$((A', A''), idx) \in \arcs$ such that $A' = A$ or $A'' = A$}{
			$\arcs \leftarrow \arcs \backslash \{(r, idx)\}$
		}
	}
 \caption{Deletion mechanism LRU}\label{alg:lru}
}
\end{algorithm*}
\endgroup
%}

\end{minipage}
} &  &  \\ \cline{2-2}
\multicolumn{1}{l|}{}                  &  & {% !TEX root = ../paper.tex

\makeatletter
\newcommand{\removelatexerror}{\let\@latex@error\@gobble}
\makeatother

\begin{minipage}{0.45\linewidth}

\begingroup
\removelatexerror% Nullify \@latex@error

\begin{algorithm*}[H]{
  \scriptsize
  \SetKwInOut{Input}{input}
  \SetKwProg{Fn}{Function}{}{}
  \SetKwFor{Loop}{Loop}{}{EndLoop}
  \Input{$\arcs, \nodes$}

	$min_a \leftarrow min\{ \freq(A) | A \in \nodes \}$\\
	$min_r \leftarrow min\{ \freq(r) | r \in \arcs \}$\\
	\uIf{$min_a > min_r$}{
		$\arcs \leftarrow \arcs \backslash \{ r \}$
	}\Else{
		$\nodes \leftarrow \nodes \backslash \{ A \}$\\
		\For{$(A', A'') \in \arcs$ such that $A' = A$ or $A'' = A$}{
			$\arcs \leftarrow \arcs \backslash \{r\}$
		}
	}
 \caption{Deletion mechanism LFU}\label{alg:lfu}
}
\end{algorithm*}
\endgroup
%}

\end{minipage}
} &  &  \\ \cline{2-2}
\multicolumn{1}{l|}{}                  &  & {% !TEX root = ../paper.tex
\makeatletter
\newcommand{\removelatexerror}{\let\@latex@error\@gobble}
\makeatother

\begin{minipage}{0.45\linewidth}
\begingroup
\removelatexerror% Nullify \@latex@error

\begin{algorithm*}[H]{
  \scriptsize
  \SetKwInOut{Input}{input}
  \SetKwProg{Fn}{Function}{}{}
  \SetKwFor{Loop}{Loop}{}{EndLoop}
  \Input{$\arcs, \nodes$}

	$min_a \leftarrow min\{ \freq(A) + L | (A, L) \in \nodes \}$\\
	$min_r \leftarrow min\{ \freq(r) + L  | (r, L) \in \arcs \}$\\
	\uIf{$min_a > min_r$}{
		$\arcs \leftarrow \arcs \backslash \{ (r, L) \}$
	}\Else{
		$\nodes \leftarrow \nodes \backslash \{ (A, L) \}$\\
		\For{$((A', A''), L) \in \arcs$ such that $A' = A$ or $A'' = A$}{
			$\arcs \leftarrow \arcs \backslash \{(r, L)\}$
		}
	}
 \caption{Deletion mechanism LFU-DA}\label{alg:lfu:da}
}
\end{algorithm*}
\endgroup
%}

\end{minipage}
} &  &  \\
                                        &                       &  &  &
\end{tabular}
\caption{Algorithms}
\label{algorithms}
\end{table*}

Consider the directly follows graph in Fig.~\ref{fig:del:strat}, where every arc is annotated with its frequency and its ``last seen index'' in parenthesis. If LRU is chosen as the deletion mechanism, then the arc representing $Send\ Bill \Rightarrow Send \ Reminder$ is deleted since this relation is the least recently observed. On the other hand, if LFU is chosen, then the relation to remove is $Send\ Bill \Rightarrow Process \ Payment$ (and similarly for LFU-DA during the first deletion), since this relation is the least frequent one.

The deletion mechanism for $S_{RC}$ imposes different challenges than the one used for $S_{PM}$. The information kept in $S_{RC}$ is necessary to compute the directly follows relations, but, if the amount of cases is unknown, then no ideal budget $B_{RC}$ can be defined. Clearly, if the end of every case is known (last event of every trace) then an entry can be eliminated from $S_{RC}$ once the corresponding case is ended. Nevertheless, other strategies can be implemented based on time or resources, e.g., a case can be deleted from $S_{RC}$ if it has been running longer than a given threshold. The size of $S_{RC}$ is denoted as $|S_{RC}|$.

Algorithm~\ref{alg:process:map} shows the procedure to construct the process map from an event stream.
The highlighted lines in Algorithm~\ref{alg:process:map} apply depending on the chosen deletion mechanism: LRU, LFU or LFU-DA (see the annotation on each highlighted line).
The algorithm starts by reading an incoming event $e$ from the event stream (method $observe$ at line 5). If the activity of $e$ is already in $\nodes$, then its frequency as well as its index (in the case of LRU) are updated (lines 6-8); otherwise, if the activity is not contained in $\nodes$ then it is necessary to check whether there is enough space to store the new activity (line 10). If the partition $S_{PM}$ is full then one of the deletion mechanisms is applied: LRU (Alg.~\ref{alg:lru}), LFU (Alg.~\ref{alg:lfu}), or LFU-DA (Alg.~\ref{alg:lfu:da}). The procedure continues by inserting the activity into $\nodes$ (lines 14-16). If the event belongs to a case $c$ that has been observed previously and the directly follows relation between the activities of the last observed event in $c$ and of the new event $e$ is present in $\arcs$, then its frequency and index are updated (lines 20-22). Otherwise, the algorithm proceeds by inserting the relation in $\arcs$ (lines 28-30), after applying the deletion mechanism if necessary (lines 24-27). The activity of the observed event is inserted into $S_{RC}$ (line 32). Finally, if the case is expired (e.g., event $e$ is the end event of the case), then it is deleted from $S_{RC}$.

Algorithms~\ref{alg:lru}-\ref{alg:lfu:da} show the deletion mechanisms. All these mechanisms work in a similar way: first, they detect the elements to delete (depending on the recency, frequency or frequency with dynamic aging) and, then, they remove them from $\nodes$ or $\arcs$. Note that if the element to be deleted is an activity, then all the relations involving that activity are removed as well.

\begin{figure*}[t!]
\centering
	\includegraphics[scale=0.4]{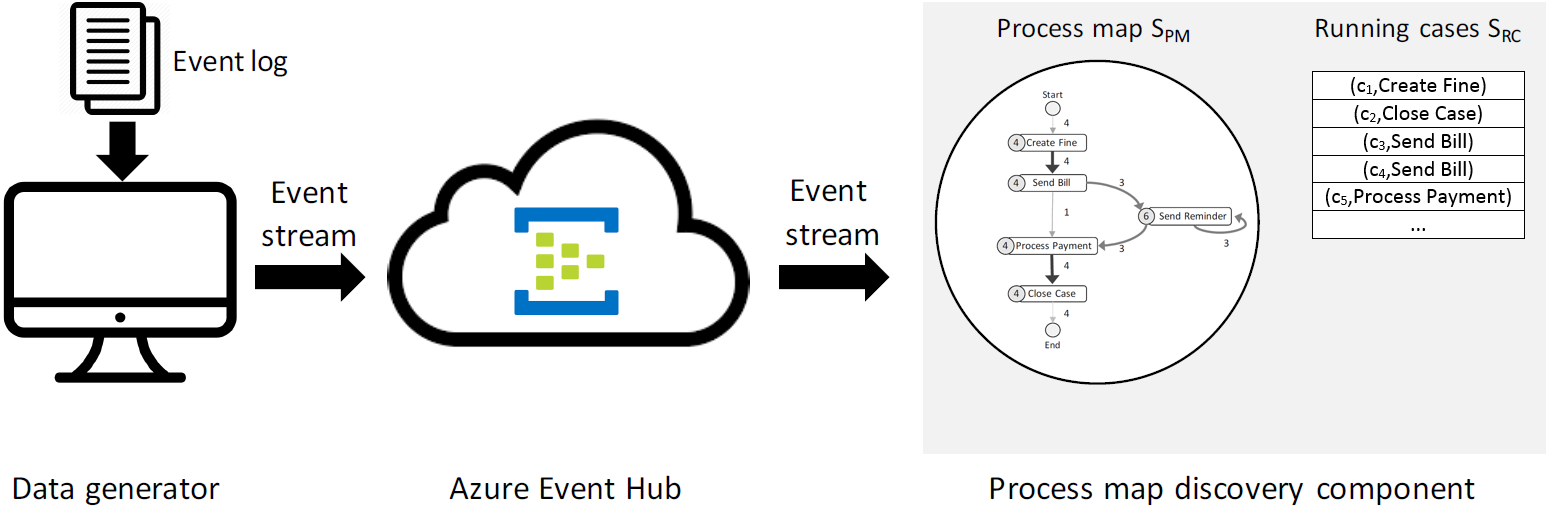}
	\caption{Evaluation setup}
	\label{fig:eval:setup}
\end{figure*}

\begin{table}[tb!]
%\vspace{-3mm}
\centering
\scriptsize
\begin{tabular}{|c|r|r|r|r|}
\hline
{\bf Dataset} & {\bf \# Traces}  & {\bf \# Events} & \thead{\bf \tiny \# Events per \\ \bf \tiny trace (Avg.)} & {\bf \# Activities}  \\ \hline
BPIC 2016 & 660,270 & 7,174,934 & 10.86 & 600  \\ \hline
BPIC 2018 & 43,809 & 2,514,266 & 57.39 & 170 \\ \hline
University log & 174,842 & 2,099,835 & 12.00 & 310 \\ \hline
%Cloud Invoice Approval & 5330 & 66074 & 12.39 & 21 \\ \hline
\end{tabular}
\caption{Datasets}
\label{table:datasets}
\end{table}

\begin{table}[tb!]
\centering
\scriptsize
%\vspace{-3mm}
\begin{tabular}{|c|r|r|r|}
\hline
{\bf  Technique }  & {\bf Activity} & {\bf Relation} & {\bf Case} \\ \hline
LCB & 3        & 4        & 4    \\ \hline
LRU & 3        & 4        & 3    \\ \hline
LFU & 2        & 3        & 3    \\ \hline
LFU-DA  & 3        & 4        & 3    \\ \hline
\end{tabular}
\caption{Words needed to store activities, relations and cases}
\label{table:cost}
\end{table}

% !TEX root = ../paper.tex
\section{Evaluation}\label{sec:evaluation} %Abel

Our approach for the discovery of directly follows graphs has been implemented in .NET and experimentally integrated into the process mining tool Minit. The tool reads an event stream from a publish-subscribe service and updates an in-memory directly follows graph whenever an event is read. In our evaluation, we use \emph{Azure event hub}\footnote{\url{https://azure.microsoft.com/en-au/services/event-hubs/}}, a highly scalable publish-subscribe service, to play the role of a reliable channel between the originator of the event stream (customer) and the process map discovery component. In order to simulate a customer, a data generator component was also implemented in .NET. The data generator reads a log and streams every event to the publish-subscribe service. The evaluation setup is depicted in Fig.~\ref{fig:eval:setup}.

\begin{figure}[t!]
	\centering
	\subfloat[BPIC 2016]{
		 \includegraphics[scale=0.47]{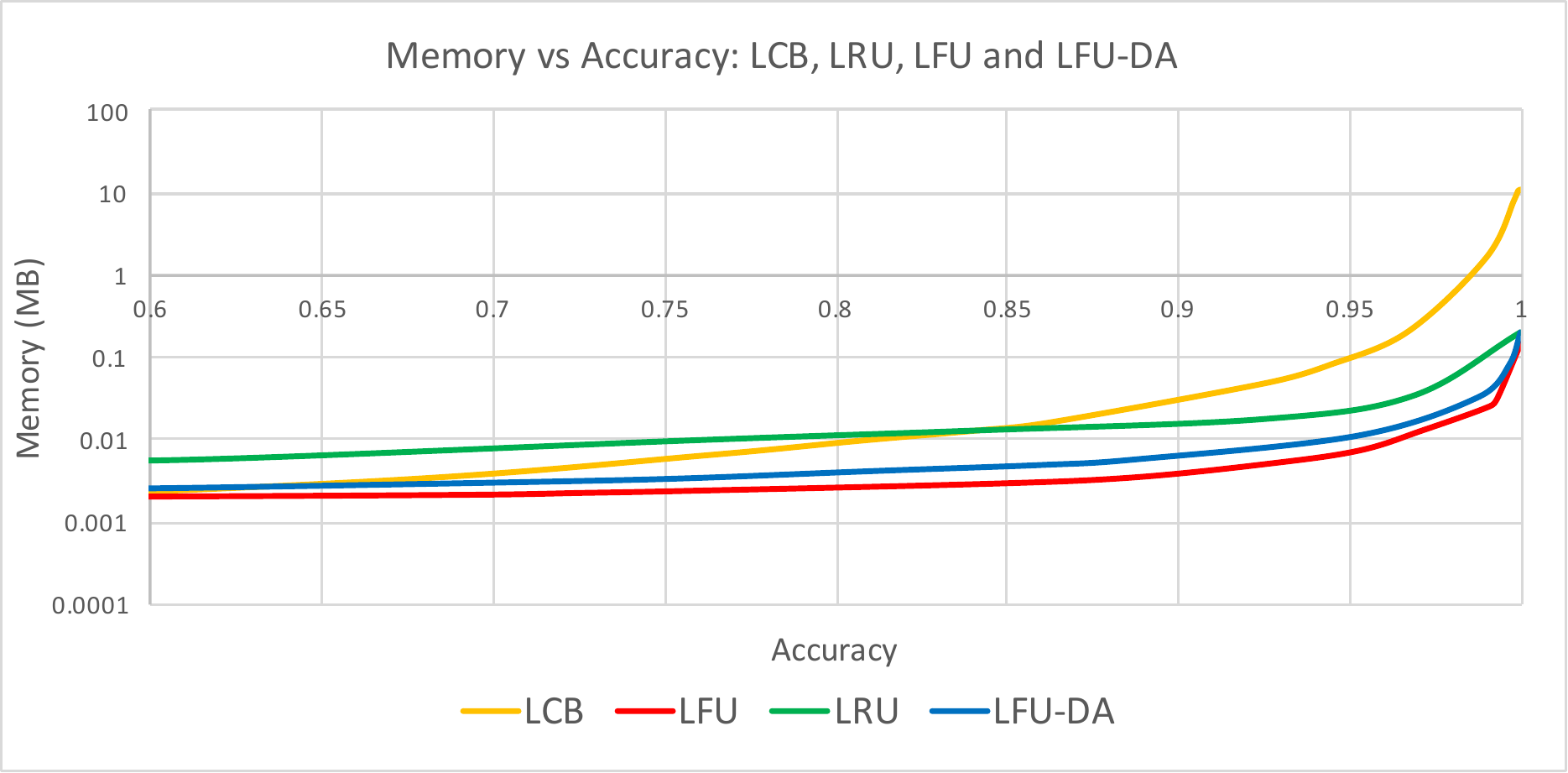}\label{fig:res1}
	}
	\hspace{3mm}
	\subfloat[BPIC 2018]{
		 \includegraphics[scale=0.555]{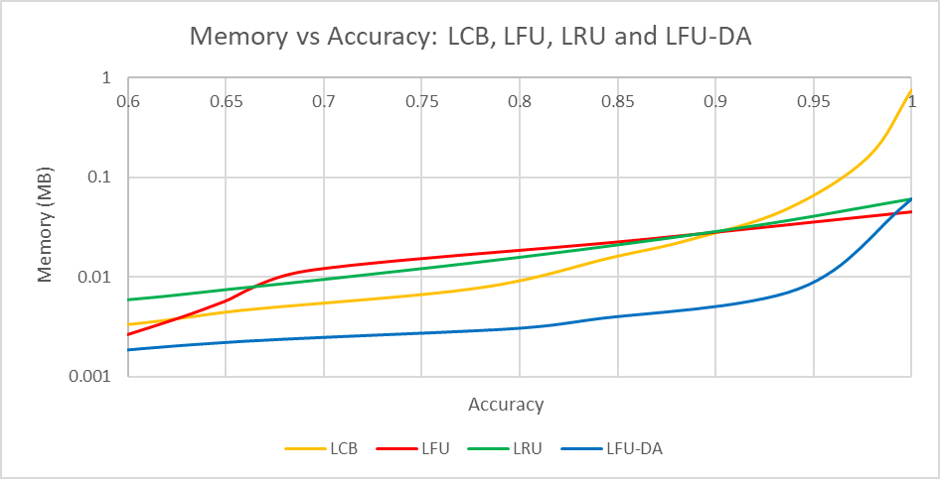}\label{fig:res2}
	}
%\subfloat[]{\includegraphics[scale=0.40]{figures/memoryAcc2.pdf}\label{fig:res2}}
\hspace{3mm}
	\subfloat[University log]{
		 \includegraphics[scale=0.48]{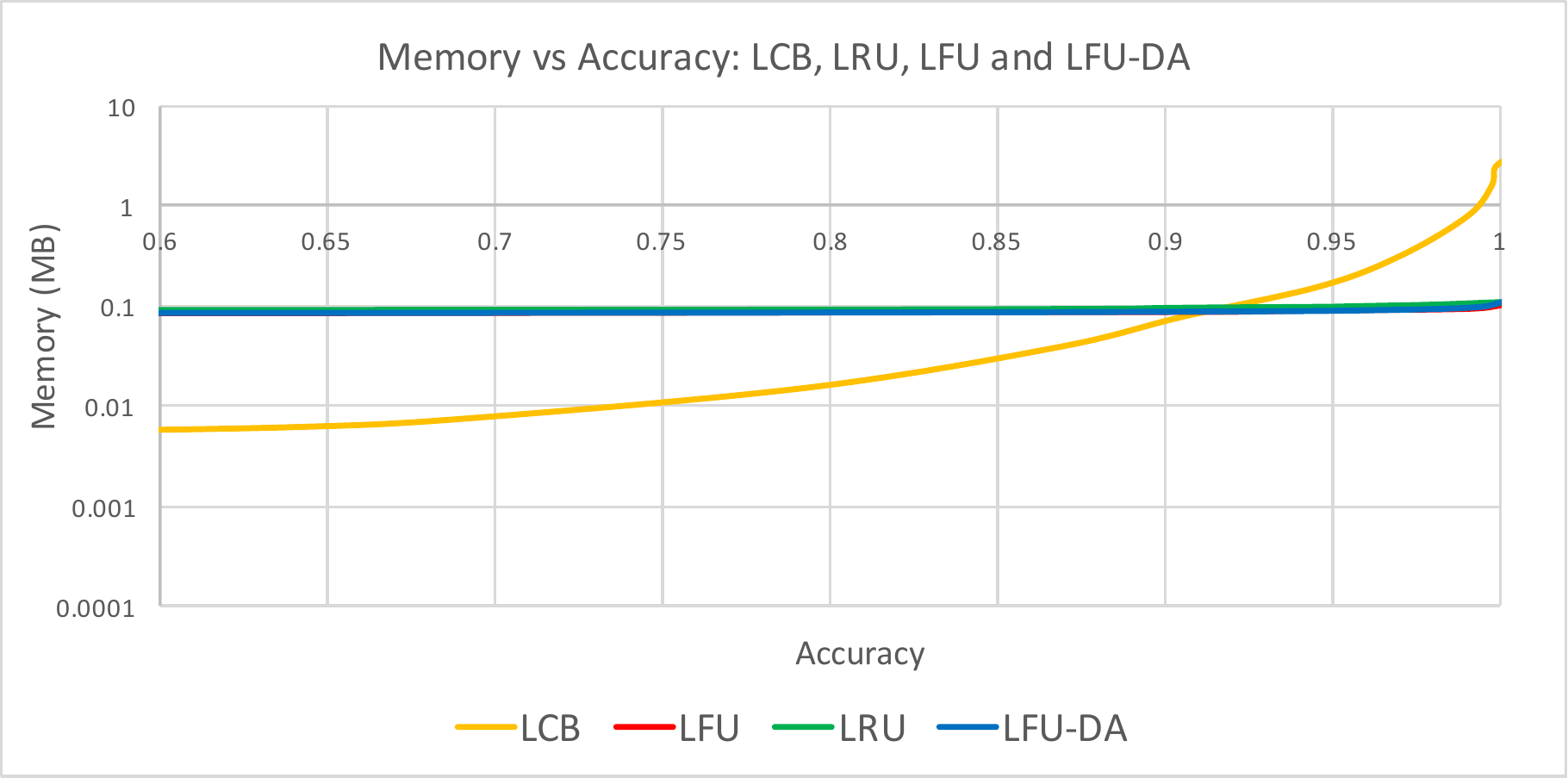}\label{fig:res3}
	}
\caption{Memory vs. Accuracy}
\label{fig:resmem}
\end{figure}

%We used three large event logs in our evaluation: BPI Challenge 2016\footnote{\url{doi:10.4121/uuid:01345ac4-7d1d-426e- 92b8-24933a079412}}, BPI Challenge 2017\footnote{\url{doi:10.4121/uuid:5f3067df-f10b-45da-b98b-86ae4c7a310b}} and a University log.
%BPI Challenge 2016 is a publicly available log pertaining to the process that customers go through in a Dutch administrative authority to manage unemployment benefits. The data was collected over a period of eight months.
%BPI Challenge 2017 is a publicly available log that describes a loan origination process of a Dutch financial institution. It contains records about loan applications filed through an online system in 2016 and their subsequent events until February 2017.
%Finally, the University log refers to a process from an accounting department of an Italian University.
%The number of traces, number of events, average number of events per trace, and number of distinct activities for each of these logs are shown in Table~\ref{table:datasets}.

\begin{figure}[t!]
	\centering
	\subfloat[BPIC 2016]{
		 \includegraphics[scale=0.477]{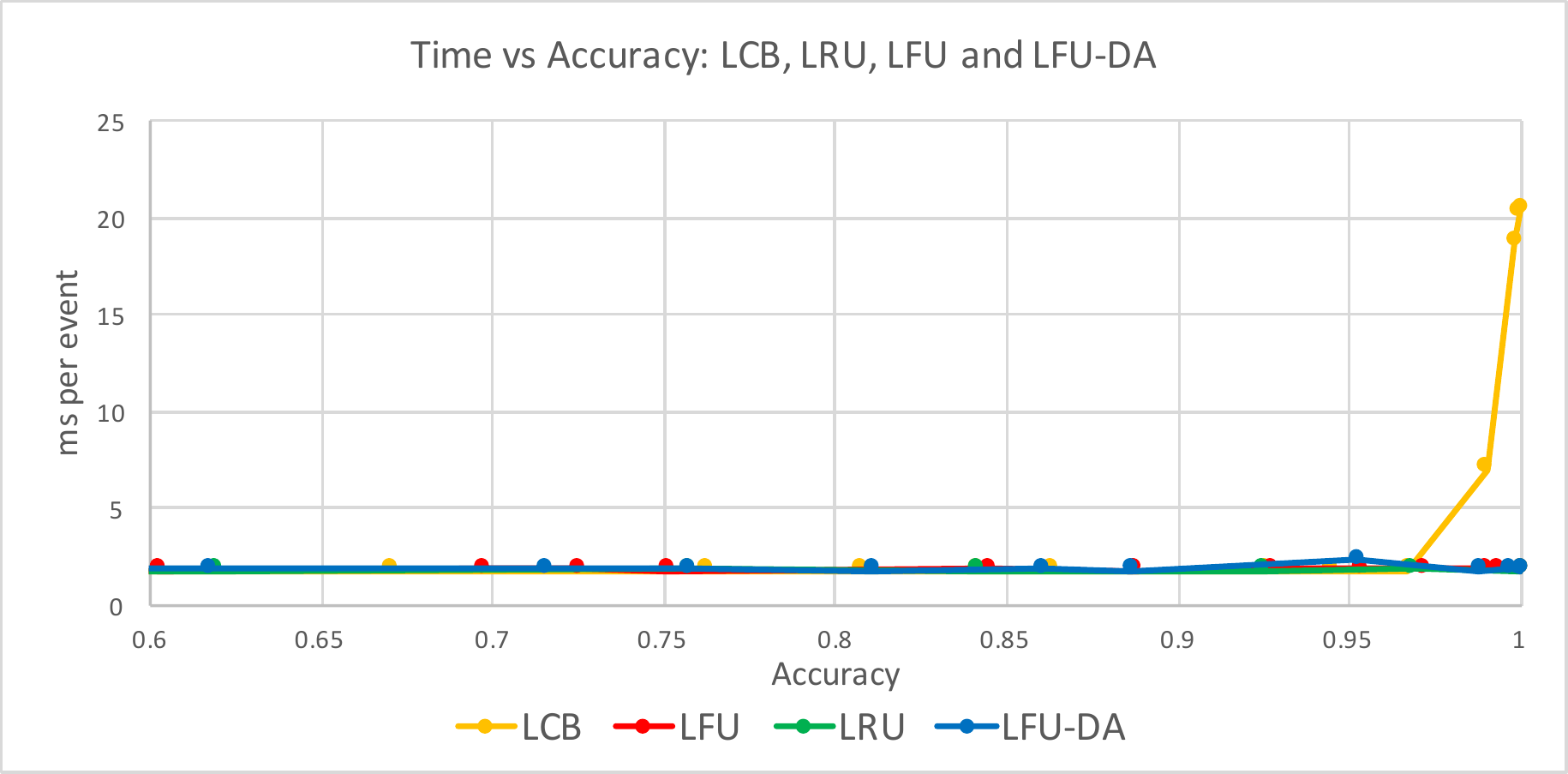}\label{fig:time:res1}
	}
	\hspace{3mm}
	\subfloat[BPIC 2018]{
		 \includegraphics[scale=0.55]{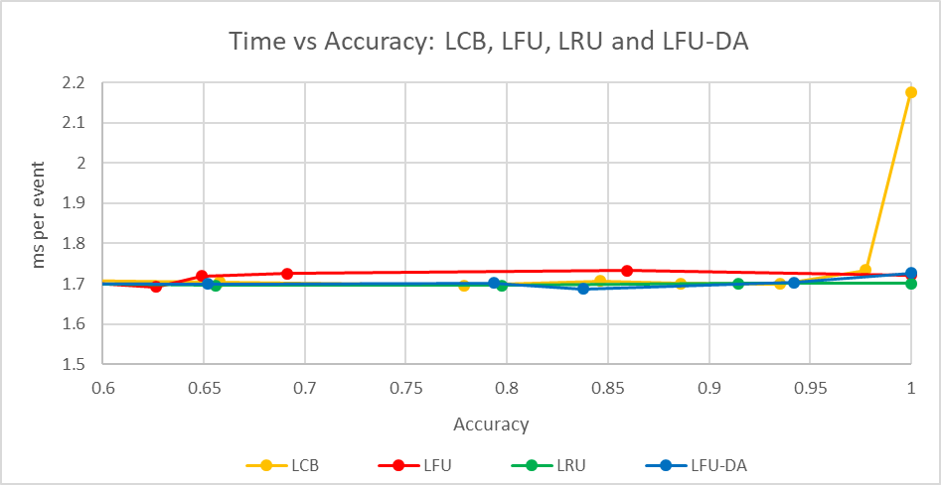}\label{fig:time:res2}
	}
%\subfloat[]{\includegraphics[scale=0.40]{figures/memoryAcc2.pdf}\label{fig:res2}}
\hspace{3mm}
	\subfloat[University log]{
		 \includegraphics[scale=0.48]{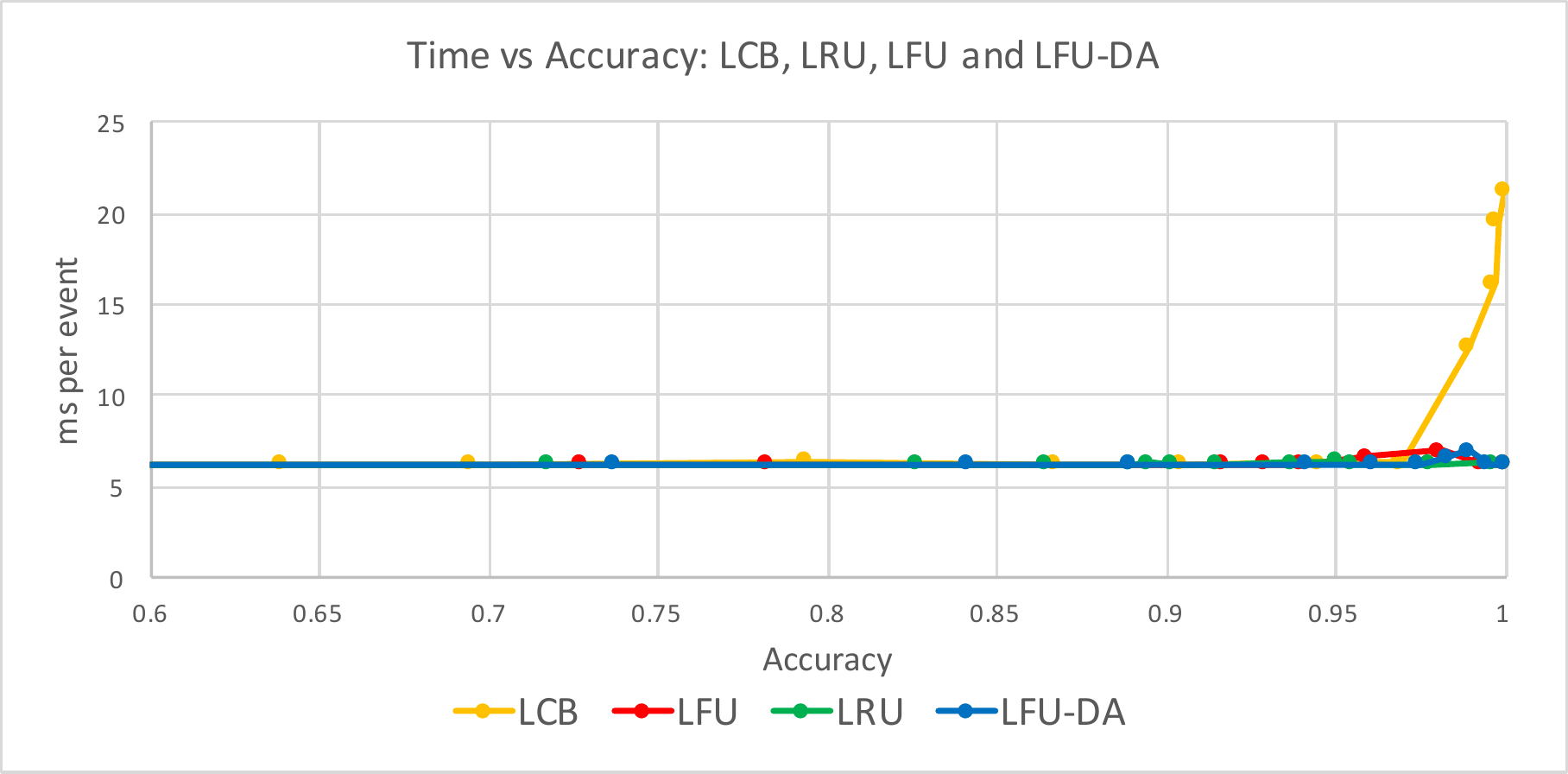}\label{fig:time:res3}
	}
%\vspace{-2mm}
\caption{Time vs. Accuracy}
\label{fig:res:time:both}
%\vspace{-4mm}
\end{figure}

We used three large event logs in our evaluation: BPI Challenge (BPIC) 2016\footnote{\url{doi:10.4121/uuid:01345ac4-7d1d-426e- 92b8-24933a079412}}, BPIC 2018\footnote{\url{doi.org/10.4121/uuid:3301445f-95e8-4ff0-98a4-901f1f204972}}, and a University log.
BPIC 2016 is a publicly available log pertaining to the process that customers go through in a Dutch administrative authority to manage unemployment benefits. The data was collected over a period of eight months.
The process considered in the BPIC 2018 covers the handling of applications for EU direct payments (basic incomes decoupled from production) for German farmers from the European Agricultural Guarantee Fund.
%The processes that govern the distribution of these funds are subject to complex regulations captured in EU and national law. The member states are required to operate an Integrated Administration and Control System (IACS), which includes IT systems to support the complex processes of subsidy distribution. The process repeats every year with minor changes due to changes in EU regulations.
%BPI Challenge 2017 is a publicly available log that describes a loan origination process of a Dutch financial institution. It contains records about loan applications filed through an online system in 2016 and their subsequent events until February 2017.
The University log refers to a process from an accounting department of an Italian University.
The number of traces, number of events, average number of events per trace, and number of distinct activities for each of these logs are shown in Table~\ref{table:datasets}.

We compare the different deletion strategies in our approach (LRU, LFU and LFU-DA) against \emph{Lossy Counting with Budget (LCB)}~\cite{BurattinSA14}, which is, to the best of our knowledge, the most efficient online process discovery technique described in the literature.
%LCB is taken as a baseline and compared against our approach using the three deletion mechanisms: LRU, LFU and LFU-DA.
The comparison of the different approaches is done in terms of \emph{time per event (ms)} and \emph{memory consumption} required to obtain different levels of \emph{accuracy}.
%Note that all the charts displayed below shows accuracy higher to 0.6, since any directly follow graph with lower accuracy would be highly a highly lossy representation of the behavior.

Accuracy evaluates the similarity between the exact directly follows graph, which is discovered from the entire event log, and the one discovered by each of the online techniques (LCB, LRU, LFU and LFU-DA). Specifically, accuracy counts the differences in frequencies between the relations in both directly follows graphs. If a relation is not present in either of the graphs, then its frequency is treated as 0; whereas, if an activity is missing in the directly follows graph discovered by an online technique (which exists in the complete graph), then this activity and all the relations it is involved in are added with a frequency of 0. % (and vice versa).
The sum of all differences in frequencies constitutes the \emph{Loss}, whereas the sum of all frequencies in the complete graph constitutes the \emph{Total frequency}. Then, the accuracy of a graph created by an online technique is computed with the following formula:

\begin{equation}
%\scriptsize	
A = 1-  \frac{Loss}{Total\ frequency}.
\end{equation}

Memory consumption shows the amount of memory used during the discovery of the directly follows graph. It is measured in terms of memory words, which refer to a fixed-size unit of memory. The number of words required by each technique to store an activity, a relation and a case is displayed in Table~\ref{table:cost}.
%\begin{wraptable}{r}{5.2cm}
The total memory consumption for a technique is computed with the formula:
\begin{equation}
%\scriptsize
	M = M(activity) \times |\nodes| + M(relation) \times |\arcs| + M(case) \times |S_{RC}|,
\end{equation}
\noindent
where $M(activity)$, $M(relation)$ and $M(case)$ are the number of words required to store activities, relations and cases (see Table~\ref{table:cost}).

%LCB also requires a notion of budget. However, differently from our approach, the budget for LCB is divided among all the internal data structures. These data structures grow as required, and the deletion mechanism is applied once the sum of the sizes of the data structures equates the budget. Instead, the budget in our techniques is used only for the storage of the directly follows graph, while the partition recording the running cases grows as required.

In Fig.~\ref{fig:resmem}, we compare LCB with the three different deletion mechanisms of our approach: LRU, LFU and LFU-DA by evaluating the memory consumption\footnote{In the plots, we report the maximal values of memory used.} required to obtain different levels of accuracy\footnote{Note that, for all the techniques, the input parameter \emph{budget} has been set in order to obtain each given value of accuracy.}. In all the approaches, we assume that the end event for each case is known. Thus, as mentioned in the previous section, once the end event of a case is read from the stream, the case is deleted from $S_{RC}$. The memory is reported in terms of MB (a word is equal to 4 bytes) and we use logarithmic scale on the \emph{Memory} axis.
%In our approach, the budget for $S_{PM}$ was increased from 0 to 12,600 objects (activities, relations and cases). The latter is the necessary budget for a lossless directly follows graph for this dataset. In the case of LCB, the budget was progressively increased from 0 to 700,000, which is the number of objects required to obtain a lossless directly follows graph.
%Note that LCB does not consider the information about the end of the cases, thus it keeps some information in memory about the running cases.
%Figure~\ref{fig:res2} shows the results for the same experiment using the dataset BPI Challenge 2017. In this case, the budget was increased from 0 to 3,500 objects for LCB, and from 0 to 200 objects for our approach.
%Finally, Fig.~\ref{fig:res3} depicts the results for the University log. In this case the budget was increased from 0 to 176,870 objects for LCB, and from 0 to 2,026 objects for our approach.

In order to obtain a lossless directly follows graph for BPIC 2016 (Fig.~\ref{fig:res1}), the memory required is LCB = 10.76 MB, LRU = 0.2 MB, LFU = 0.15 MB, and LFU-DA = 0.2 MB. Therefore, LCB requires a considerably larger amount of memory (two orders of magnitude higher) than the three variants of our approach.
The amount of memory used by LCB starts to be higher than the amount of memory required by LFU and LFU-DA when the accuracy exceeds 0.65, while in the case of LRU, it happens when the accuracy exceeds 0.85.
The memory required for a lossless directly follows graph for BPIC 2018 (Fig.~\ref{fig:res2}) is LCB = 0.76 MB, LRU = 0.07 MB, LFU = 0.06 MB, and LFU-DA = 0.07 MB. Also in this case, LCB requires a larger amount of memory (one order of magnitude higher) than our approach. In this case, the amount of memory used by LCB is always higher than the amount of memory required by LFU-DA, whereas the amount of memory used by LCB starts to be higher than the amount of memory required by LRU and LFU when the accuracy exceeds 0.9.
The results for the University log (Fig.~\ref{fig:res3}) show that LCB requires a larger amount of memory for an accuracy higher than 0.9, while the amount of memory is very similar for all variants of our approach. The amount of memory for a lossless directly follows graph is LCB = 2.82 MB, LRU = 0.10 MB, LFU = 0.10 MB, and LFU-DA = 0.10 MB, and LCB uses one order of magnitude more memory than our approach.

Figure~\ref{fig:res:time:both} reports time performance (ms per event) vs.\ accuracy. For BPIC 2016 (Fig.~\ref{fig:time:res1}), LCB takes significantly longer when the accuracy approaches 1.0, while our approach performs in a similar way across its different variants. Figure~\ref{fig:time:res2} shows the results for BPIC 2018. We can see that, for this log, LRU, LFU and LFU-DA take around 1.7 ms per event when the accuracy is 1.0, while LCB requires 2.19 ms.
%Although all techniques perform similarly in this case, LCB requires more time on average and takes 1.72 ms per event when the accuracy is 1.0, while LRU, LFU and LFU-DA take 1.53 ms.
The results for the University log are displayed in Fig.~\ref{fig:time:res3}. Here, LRU, LFU and LFU-DA take 6.22 ms per event when the accuracy is 1.0, while LCB requires 21.29 ms. The variations in time in Fig.~\ref{fig:res:time:both} show that, when the end event of each case is known, the deletion mechanisms in all the variants of our approach are more efficient than the one in LCB. %Although in most of the cases the variants of our approach did not vary much in time, LRU applied over the University log displays some spikes.

%Finally, we considered the case when the end event of a case is unknown and no case is removed. In this case, LCB and all variants of our approach have similar performance in terms of memory consumption and time per event. However, using LCB, the user does not have any control over the accuracy of the resulting process map. Our approach, instead, provides a clear method to identify the memory budget needed in order to achieve a given level of accuracy.
%intuitive way for managing the memory used for the construction of the directly follows graph.
%As an example see Fig.~\ref{fig:res3} and Fig.~\ref{fig:res4}, they contain the results for LCB and LFU-DA (with logarithmic scale in the axis ``Memory'') for BPI Challenge 2016 and BPI Challenge 2017, respectively.

%%\begin{wrapfigure}{l}{6cm}
%\begin{figure}
%\centering
%\includegraphics[scale=0.40]{figures/memoryAcc3Log.pdf}
%%\subfloat[]{\includegraphics[scale=0.32]{figures/memoryAcc4.pdf}\label{fig:res4}}
%\caption{Memory vs. Accuracy for BPI Challenge 2016 -- LCB and LFU-DA}
%\label{fig:res3}
%%\end{wrapfigure}
%\end{figure}
%
%%\begin{wrapfigure}{l}{6cm}
%\begin{figure}
%\centering
%\includegraphics[scale=0.40]{figures/memoryAcc4Log.pdf}
%%\subfloat[]{\includegraphics[scale=0.32]{figures/memoryAcc4.pdf}\label{fig:res4}}
%\caption{Memory vs. Accuracy for BPI Challenge 2017 -- LCB and LFU-DA}
%\label{fig:res4}
%%\end{wrapfigure}
%\end{figure}

% !TEX root = ../paper.tex
\section{Conclusion}\label{sec:conclusion} %Marlon
This paper proposes an approach for the online discovery of process maps from a stream of events.
%Process maps are a common process representation supported by commercial process mining tools, and also employed as an internal structure within more-advanced process discovery techniques such as the Inductive Miner and the Heuristics Miner.
The approach relies on three cache replacement policies to maintain an up-to-date in-memory representation of the process map, which is updated whenever a new process event or behavioral relation between process events is observed from the stream.
%The approach has been implemented in .NET, experimentally integrated with the Minit process mining tool and compared to the state of the art in online process discovery using real-life datasets.
The evaluation on real-life datasets shows that our approach outperforms the state of the art, in terms amount of memory and time per event. The performance is comparable if the last event of each case is not known. In addition, the approach provides a clear way to control the accuracy of the resulting process map.

As future work, we plan to use cache replacement policies for the online discovery of other types of process models such as declarative models and social networks of the interactions among process participants. We also plan to discover process maps that consider other metrics along with the frequency of the directly follows relations, such as the time difference between adjacent events, as a means to evaluate the performance of the underlying business process.

%\vspace{3mm}
%\textbf{Acknowledgments.} This work was partly conducted while the first author did an internship at MinitLabs. We wish to thank Michal Rosik from MinitLabs for his valuable feedback. This research is funded by the Australian Research Council (DP150103356) and the Estonian Research Council (IUT20-55).

\bibliographystyle{ACM-Reference-Format}
\bibliography{bibliography}


\begin{thebibliography}{00}

%%% the Web address.

\ifx \showCODEN    \undefined \def \showCODEN     #1{\unskip}     \fi
\ifx \showDOI      \undefined \def \showDOI       #1{#1}\fi
\ifx \showISBNx    \undefined \def \showISBNx     #1{\unskip}     \fi
\ifx \showISBNxiii \undefined \def \showISBNxiii  #1{\unskip}     \fi
\ifx \showISSN     \undefined \def \showISSN      #1{\unskip}     \fi
\ifx \showLCCN     \undefined \def \showLCCN      #1{\unskip}     \fi
\ifx \shownote     \undefined \def \shownote      #1{#1}          \fi
\ifx \showarticletitle \undefined \def \showarticletitle #1{#1}   \fi
\ifx \showURL      \undefined \def \showURL       {\relax}        \fi
% invisible to TeX
\providecommand\bibfield[2]{#2}
\providecommand\bibinfo[2]{#2}
\providecommand\natexlab[1]{#1}
\providecommand\showeprint[2][]{arXiv:#2}

\bibitem[\protect\citeauthoryear{Agrawal, Gunopulos, and Leymann}{Agrawal
  et~al\mbox{.}}{1998}]%
        {Agrawal1998:Mining}
\bibfield{author}{\bibinfo{person}{Rakesh Agrawal}, \bibinfo{person}{Dimitrios
  Gunopulos}, {and} \bibinfo{person}{Frank Leymann}.}
  \bibinfo{year}{1998}\natexlab{}.
\newblock \showarticletitle{Mining Process Models from Workflow Logs}. In
  \bibinfo{booktitle}{{\em Advances in Database Technology - EDBT'98, 6th
  International Conference on Extending Database Technology, Valencia, Spain,
  March 23-27, 1998, Proceedings}}. \bibinfo{pages}{469--483}.
\newblock
\showDOI{%
\url{https://doi.org/10.1007/BFb0101003}}


\bibitem[\protect\citeauthoryear{Arlitt, Cherkasova, Dilley, Friedrich, and
  Jin}{Arlitt et~al\mbox{.}}{2000}]%
        {ArlittCDFJ00}
\bibfield{author}{\bibinfo{person}{Martin~F. Arlitt}, \bibinfo{person}{Ludmila
  Cherkasova}, \bibinfo{person}{John Dilley}, \bibinfo{person}{Rich Friedrich},
  {and} \bibinfo{person}{Tai Jin}.} \bibinfo{year}{2000}\natexlab{}.
\newblock \showarticletitle{Evaluating content management techniques for Web
  proxy caches}.
\newblock \bibinfo{journal}{{\em {SIGMETRICS} Performance Evaluation Review\/}}
  \bibinfo{volume}{27}, \bibinfo{number}{4} (\bibinfo{year}{2000}),
  \bibinfo{pages}{3--11}.
\newblock
\showDOI{%
\url{https://doi.org/10.1145/346000.346003}}


\bibitem[\protect\citeauthoryear{Augusto, Conforti, Dumas, {La Rosa}, Maggi,
  Marrella, Mecella, and Soo}{Augusto et~al\mbox{.}}{2017}]%
        {Augusto2017Survey}
\bibfield{author}{\bibinfo{person}{Adriano Augusto}, \bibinfo{person}{Raffaele
  Conforti}, \bibinfo{person}{Marlon Dumas}, \bibinfo{person}{Marcello {La
  Rosa}}, \bibinfo{person}{Fabrizio~Maria Maggi}, \bibinfo{person}{Andrea
  Marrella}, \bibinfo{person}{Massimo Mecella}, {and} \bibinfo{person}{Allar
  Soo}.} \bibinfo{year}{2017}\natexlab{}.
\newblock \showarticletitle{Automated Discovery of Process Models from Event
  Logs: Review and Benchmark}.
\newblock \bibinfo{journal}{{\em ArXiv CoRR\/}} (\bibinfo{year}{2017}).
\newblock


\bibitem[\protect\citeauthoryear{Bose}{Bose}{2012}]%
        {jcthesis}
\bibfield{author}{\bibinfo{person}{R.P. Jagadeesh~Chandra Bose}.}
  \bibinfo{year}{2012}\natexlab{}.
\newblock {\em \bibinfo{title}{{Process Mining in the Large: Preprocessing,
  Discovery, and Diagnostics}}}.
\newblock \bibinfo{thesistype}{Ph.D. Dissertation}. \bibinfo{school}{Eindhoven
  University of Technology}.
\newblock


\bibitem[\protect\citeauthoryear{Burattin, Cimitile, and Maggi}{Burattin
  et~al\mbox{.}}{2014a}]%
        {BurattinCM14}
\bibfield{author}{\bibinfo{person}{Andrea Burattin}, \bibinfo{person}{Marta
  Cimitile}, {and} \bibinfo{person}{Fabrizio~Maria Maggi}.}
  \bibinfo{year}{2014}\natexlab{a}.
\newblock \showarticletitle{Lights, Camera, Action! Business Process Movies for
  Online Process Discovery}. In \bibinfo{booktitle}{{\em Business Process
  Management Workshops - {BPM} 2014 International Workshops, Eindhoven, The
  Netherlands, September 7-8, 2014, Revised Papers}}.
  \bibinfo{pages}{408--419}.
\newblock
\showDOI{%
\url{https://doi.org/10.1007/978-3-319-15895-2_34}}


\bibitem[\protect\citeauthoryear{Burattin, Cimitile, Maggi, and
  Sperduti}{Burattin et~al\mbox{.}}{2015}]%
        {DBLP:journals/tsc/BurattinCMS15}
\bibfield{author}{\bibinfo{person}{Andrea Burattin}, \bibinfo{person}{Marta
  Cimitile}, \bibinfo{person}{Fabrizio~Maria Maggi}, {and}
  \bibinfo{person}{Alessandro Sperduti}.} \bibinfo{year}{2015}\natexlab{}.
\newblock \showarticletitle{Online Discovery of Declarative Process Models from
  Event Streams}.
\newblock \bibinfo{journal}{{\em {IEEE} Trans. Services Computing\/}}
  \bibinfo{volume}{8}, \bibinfo{number}{6} (\bibinfo{year}{2015}),
  \bibinfo{pages}{833--846}.
\newblock
\showDOI{%
\url{https://doi.org/10.1109/TSC.2015.2459703}}


\bibitem[\protect\citeauthoryear{Burattin, Sperduti, and van~der
  Aalst}{Burattin et~al\mbox{.}}{2012}]%
        {Burattin2012b}
\bibfield{author}{\bibinfo{person}{Andrea Burattin},
  \bibinfo{person}{Alessandro Sperduti}, {and} \bibinfo{person}{Wil M.~P.
  van~der Aalst}.} \bibinfo{year}{2012}\natexlab{}.
\newblock \showarticletitle{{Heuristics Miners for Streaming Event Data}}.
\newblock \bibinfo{journal}{{\em ArXiv CoRR\/}} (\bibinfo{date}{Dec.}
  \bibinfo{year}{2012}).
\newblock
\showeprint[arxiv]{1212.6383}
\showURL{%
\url{http://arxiv.org/abs/1212.6383}}


\bibitem[\protect\citeauthoryear{Burattin, Sperduti, and van~der
  Aalst}{Burattin et~al\mbox{.}}{2014b}]%
        {BurattinSA14}
\bibfield{author}{\bibinfo{person}{Andrea Burattin},
  \bibinfo{person}{Alessandro Sperduti}, {and} \bibinfo{person}{Wil M.~P.
  van~der Aalst}.} \bibinfo{year}{2014}\natexlab{b}.
\newblock \showarticletitle{Control-flow discovery from event streams}. In
  \bibinfo{booktitle}{{\em Proceedings of the {IEEE} Congress on Evolutionary
  Computation, {CEC} 2014, Beijing, China, July 6-11, 2014}}.
  \bibinfo{pages}{2420--2427}.
\newblock
\showDOI{%
\url{https://doi.org/10.1109/CEC.2014.6900341}}


\bibitem[\protect\citeauthoryear{{Da San Martino}, Navarin, and Sperduti}{{Da
  San Martino} et~al\mbox{.}}{2013}]%
        {DaSanMartino2012}
\bibfield{author}{\bibinfo{person}{Giovanni {Da San Martino}},
  \bibinfo{person}{Nicol{\`{o}} Navarin}, {and} \bibinfo{person}{Alessandro
  Sperduti}.} \bibinfo{year}{2013}\natexlab{}.
\newblock \showarticletitle{A Lossy Counting Based Approach for Learning on
  Streams of Graphs on a Budget}. In \bibinfo{booktitle}{{\em {IJCAI} 2013,
  Proceedings of the 23rd International Joint Conference on Artificial
  Intelligence, Beijing, China, August 3-9, 2013}}.
  \bibinfo{pages}{1294--1301}.
\newblock
\showURL{%
\url{http://www.aaai.org/ocs/index.php/IJCAI/IJCAI13/paper/view/6699}}


\bibitem[\protect\citeauthoryear{{De Weerdt}, {De Backer}, Vanthienen, and
  Baesens}{{De Weerdt} et~al\mbox{.}}{2012}]%
        {WeerdtBVB12}
\bibfield{author}{\bibinfo{person}{Jochen {De Weerdt}}, \bibinfo{person}{Manu
  {De Backer}}, \bibinfo{person}{Jan Vanthienen}, {and} \bibinfo{person}{Bart
  Baesens}.} \bibinfo{year}{2012}\natexlab{}.
\newblock \showarticletitle{A multi-dimensional quality assessment of
  state-of-the-art process discovery algorithms using real-life event logs}.
\newblock \bibinfo{journal}{{\em Inf. Syst.\/}} \bibinfo{volume}{37},
  \bibinfo{number}{7} (\bibinfo{year}{2012}), \bibinfo{pages}{654--676}.
\newblock
\showDOI{%
\url{https://doi.org/10.1016/j.is.2012.02.004}}


\bibitem[\protect\citeauthoryear{Hassani, Siccha, Richter, and Seidl}{Hassani
  et~al\mbox{.}}{2015}]%
        {hassaniSRS2015}
\bibfield{author}{\bibinfo{person}{Marwan Hassani}, \bibinfo{person}{Sergio
  Siccha}, \bibinfo{person}{Florian Richter}, {and} \bibinfo{person}{Thomas
  Seidl}.} \bibinfo{year}{2015}\natexlab{}.
\newblock \showarticletitle{Efficient Process Discovery From Event Streams
  Using Sequential Pattern Mining}. In \bibinfo{booktitle}{{\em {IEEE}
  Symposium Series on Computational Intelligence, {SSCI} 2015, Cape Town, South
  Africa, December 7-10, 2015}}. \bibinfo{pages}{1366--1373}.
\newblock
\showDOI{%
\url{https://doi.org/10.1109/SSCI.2015.195}}


\bibitem[\protect\citeauthoryear{Kindler, Rubin, and Sch{\"{a}}fer}{Kindler
  et~al\mbox{.}}{2005}]%
        {Kindler2005a}
\bibfield{author}{\bibinfo{person}{Ekkart Kindler},
  \bibinfo{person}{Vladimir~A. Rubin}, {and} \bibinfo{person}{Wilhelm
  Sch{\"{a}}fer}.} \bibinfo{year}{2005}\natexlab{}.
\newblock \showarticletitle{Incremental Workflow Mining Based on Document
  Versioning Information}. In \bibinfo{booktitle}{{\em Unifying the Software
  Process Spectrum, International Software Process Workshop, {SPW} 2005,
  Beijing, China, May 25-27, 2005, Revised Selected Papers}}.
  \bibinfo{pages}{287--301}.
\newblock
\showDOI{%
\url{https://doi.org/10.1007/11608035_25}}


\bibitem[\protect\citeauthoryear{Kindler, Rubin, and Sch{\"{a}}fer}{Kindler
  et~al\mbox{.}}{2006}]%
        {Kindler2006c}
\bibfield{author}{\bibinfo{person}{Ekkart Kindler},
  \bibinfo{person}{Vladimir~A. Rubin}, {and} \bibinfo{person}{Wilhelm
  Sch{\"{a}}fer}.} \bibinfo{year}{2006}\natexlab{}.
\newblock \showarticletitle{Incremental Workflow Mining for Process
  Flexibility}. In \bibinfo{booktitle}{{\em Proceedings of the CAISE*06
  Workshop on Business Process Modelling, Development, and Support {BPMDS} '06,
  Luxemburg, June 5-9, 2006}}.
\newblock
\showURL{%
\url{http://ceur-ws.org/Vol-236/paper12.pdf}}


\bibitem[\protect\citeauthoryear{Maaradji, Dumas, {La Rosa}, and
  Ostovar}{Maaradji et~al\mbox{.}}{2015}]%
        {DBLP:conf/bpm/MaaradjiDRO15}
\bibfield{author}{\bibinfo{person}{Abderrahmane Maaradji},
  \bibinfo{person}{Marlon Dumas}, \bibinfo{person}{Marcello {La Rosa}}, {and}
  \bibinfo{person}{Alireza Ostovar}.} \bibinfo{year}{2015}\natexlab{}.
\newblock \showarticletitle{Fast and Accurate Business Process Drift
  Detection}. In \bibinfo{booktitle}{{\em Business Process Management - 13th
  International Conference, {BPM} 2015, Innsbruck, Austria, August 31 -
  September 3, 2015, Proceedings}}. \bibinfo{pages}{406--422}.
\newblock
\showDOI{%
\url{https://doi.org/10.1007/978-3-319-23063-4_27}}


\bibitem[\protect\citeauthoryear{Maggi, Burattin, Cimitile, and Sperduti}{Maggi
  et~al\mbox{.}}{2013}]%
        {DBLP:conf/otm/MaggiBCS13}
\bibfield{author}{\bibinfo{person}{Fabrizio~Maria Maggi},
  \bibinfo{person}{Andrea Burattin}, \bibinfo{person}{Marta Cimitile}, {and}
  \bibinfo{person}{Alessandro Sperduti}.} \bibinfo{year}{2013}\natexlab{}.
\newblock \showarticletitle{Online Process Discovery to Detect Concept Drifts
  in LTL-Based Declarative Process Models}. In \bibinfo{booktitle}{{\em On the
  Move to Meaningful Internet Systems: {OTM} 2013 Conferences - Confederated
  International Conferences: CoopIS, DOA-Trusted Cloud, and {ODBASE} 2013,
  Graz, Austria, September 9-13, 2013. Proceedings}}. \bibinfo{pages}{94--111}.
\newblock
\showDOI{%
\url{https://doi.org/10.1007/978-3-642-41030-7_7}}


\bibitem[\protect\citeauthoryear{Maggi, Montali, and van~der Aalst}{Maggi
  et~al\mbox{.}}{2012}]%
        {DBLP:conf/fase/MaggiMA12}
\bibfield{author}{\bibinfo{person}{Fabrizio~Maria Maggi},
  \bibinfo{person}{Marco Montali}, {and} \bibinfo{person}{Wil M.~P. van~der
  Aalst}.} \bibinfo{year}{2012}\natexlab{}.
\newblock \showarticletitle{An Operational Decision Support Framework for
  Monitoring Business Constraints}. In \bibinfo{booktitle}{{\em Fundamental
  Approaches to Software Engineering - 15th International Conference, {FASE}
  2012, Held as Part of the European Joint Conferences on Theory and Practice
  of Software, {ETAPS} 2012, Tallinn, Estonia, March 24 - April 1, 2012.
  Proceedings}}. \bibinfo{pages}{146--162}.
\newblock


\bibitem[\protect\citeauthoryear{Maggi and Westergaard}{Maggi and
  Westergaard}{2017}]%
        {DBLP:journals/eis/MaggiW17}
\bibfield{author}{\bibinfo{person}{Fabrizio~Maria Maggi} {and}
  \bibinfo{person}{Michael Westergaard}.} \bibinfo{year}{2017}\natexlab{}.
\newblock \showarticletitle{Designing software for operational decision support
  through coloured Petri nets}.
\newblock \bibinfo{journal}{{\em Enterprise {IS}\/}} \bibinfo{volume}{11},
  \bibinfo{number}{5} (\bibinfo{year}{2017}), \bibinfo{pages}{576--596}.
\newblock


\bibitem[\protect\citeauthoryear{Manku and Motwani}{Manku and Motwani}{2012}]%
        {MankuM02}
\bibfield{author}{\bibinfo{person}{Gurmeet~Singh Manku} {and}
  \bibinfo{person}{Rajeev Motwani}.} \bibinfo{year}{2012}\natexlab{}.
\newblock \showarticletitle{Approximate Frequency Counts over Data Streams}.
\newblock \bibinfo{journal}{{\em {PVLDB}\/}} \bibinfo{volume}{5},
  \bibinfo{number}{12} (\bibinfo{year}{2012}), \bibinfo{pages}{1699}.
\newblock
\showURL{%
\url{http://vldb.org/pvldb/vol5/p1699_gurmeetsinghmanku_vldb2012.pdf}}


\bibitem[\protect\citeauthoryear{Ostovar, Maaradji, {La Rosa}, and ter
  Hofstede}{Ostovar et~al\mbox{.}}{2017}]%
        {DBLP:conf/caise/OstovarMRH17}
\bibfield{author}{\bibinfo{person}{Alireza Ostovar},
  \bibinfo{person}{Abderrahmane Maaradji}, \bibinfo{person}{Marcello {La
  Rosa}}, {and} \bibinfo{person}{Arthur H.~M. ter Hofstede}.}
  \bibinfo{year}{2017}\natexlab{}.
\newblock \showarticletitle{Characterizing Drift from Event Streams of Business
  Processes}. In \bibinfo{booktitle}{{\em Advanced Information Systems
  Engineering - 29th International Conference, CAiSE 2017, Essen, Germany, June
  12-16, 2017, Proceedings}}. \bibinfo{pages}{210--228}.
\newblock
\showDOI{%
\url{https://doi.org/10.1007/978-3-319-59536-8_14}}


\bibitem[\protect\citeauthoryear{Ostovar, Maaradji, {La Rosa}, ter Hofstede,
  and van Dongen}{Ostovar et~al\mbox{.}}{2016}]%
        {DBLP:conf/er/OstovarMRHD16}
\bibfield{author}{\bibinfo{person}{Alireza Ostovar},
  \bibinfo{person}{Abderrahmane Maaradji}, \bibinfo{person}{Marcello {La
  Rosa}}, \bibinfo{person}{Arthur H.~M. ter Hofstede}, {and}
  \bibinfo{person}{Boudewijn~F. van Dongen}.} \bibinfo{year}{2016}\natexlab{}.
\newblock \showarticletitle{Detecting Drift from Event Streams of Unpredictable
  Business Processes}. In \bibinfo{booktitle}{{\em Conceptual Modeling - 35th
  International Conference, {ER} 2016, Gifu, Japan, November 14-17, 2016,
  Proceedings}}. \bibinfo{pages}{330--346}.
\newblock
\showDOI{%
\url{https://doi.org/10.1007/978-3-319-46397-1_26}}


\bibitem[\protect\citeauthoryear{Pesic, Schonenberg, and van~der Aalst}{Pesic
  et~al\mbox{.}}{2007}]%
        {Pesic2007}
\bibfield{author}{\bibinfo{person}{Maja Pesic}, \bibinfo{person}{Helen
  Schonenberg}, {and} \bibinfo{person}{Wil M.~P. van~der Aalst}.}
  \bibinfo{year}{2007}\natexlab{}.
\newblock \showarticletitle{{DECLARE:} Full Support for Loosely-Structured
  Processes}. In \bibinfo{booktitle}{{\em 11th {IEEE} International Enterprise
  Distributed Object Computing Conference {(EDOC} 2007), 15-19 October 2007,
  Annapolis, Maryland, {USA}}}. \bibinfo{pages}{287--300}.
\newblock
\showDOI{%
\url{https://doi.org/10.1109/EDOC.2007.14}}


\bibitem[\protect\citeauthoryear{Sharp and McDermott}{Sharp and
  McDermott}{2008}]%
        {Sharp2008}
\bibfield{author}{\bibinfo{person}{Alec Sharp} {and} \bibinfo{person}{Patrick
  McDermott}.} \bibinfo{year}{2008}\natexlab{}.
\newblock \bibinfo{booktitle}{{\em {Workflow Modeling: Tools for Process
  Improvement and Application Development}}}.
\newblock \bibinfo{publisher}{Artech House Publishers}. 449 pages.
\newblock
\showISBNx{1596931922}


\bibitem[\protect\citeauthoryear{van~der Aalst}{van~der Aalst}{2015}]%
        {process-mining-book-2011}
\bibfield{author}{\bibinfo{person}{Wil M.~P. van~der Aalst}.}
  \bibinfo{year}{2015}\natexlab{}.
\newblock \bibinfo{booktitle}{{\em Process Mining: Data Science in Action,
  second edition}}.
\newblock \bibinfo{publisher}{Springer}.
\newblock
\showISBNx{3642193447, 9783642193446}


\bibitem[\protect\citeauthoryear{van Zelst, van Dongen, and van~der Aalst}{van
  Zelst et~al\mbox{.}}{2018}]%
        {vanZelstDA2017}
\bibfield{author}{\bibinfo{person}{S.~J. van Zelst}, \bibinfo{person}{B.~F. van
  Dongen}, {and} \bibinfo{person}{W.~M.~P. van~der Aalst}.}
  \bibinfo{year}{2018}\natexlab{}.
\newblock \showarticletitle{Event stream-based process discovery using abstract
  representations}.
\newblock \bibinfo{journal}{{\em Knowledge and Information Systems\/}}
  \bibinfo{volume}{54}, \bibinfo{number}{2} (\bibinfo{year}{2018}),
  \bibinfo{pages}{407--435}.
\newblock
\showDOI{%
\url{https://doi.org/10.1007/s10115-017-1060-2}}


\bibitem[\protect\citeauthoryear{Weijters, van~der Aalst, and {Alves De
  Medeiros}}{Weijters et~al\mbox{.}}{2006}]%
        {WeijtersAM06}
\bibfield{author}{\bibinfo{person}{Ton Weijters}, \bibinfo{person}{Wil M.~P.
  van~der Aalst}, {and} \bibinfo{person}{A.~K. {Alves De Medeiros}}.}
  \bibinfo{year}{2006}\natexlab{}.
\newblock \showarticletitle{Process mining with the heuristics
  miner-algorithm}.
\newblock \bibinfo{journal}{{\em TU/e, Tech. Rep.\/}}  \bibinfo{volume}{166}
  (\bibinfo{year}{2006}), \bibinfo{pages}{1--34}.
\newblock


\bibitem[\protect\citeauthoryear{Westergaard and Maggi}{Westergaard and
  Maggi}{2011}]%
        {DBLP:conf/apn/WestergaardM11}
\bibfield{author}{\bibinfo{person}{Michael Westergaard} {and}
  \bibinfo{person}{Fabrizio~Maria Maggi}.} \bibinfo{year}{2011}\natexlab{}.
\newblock \showarticletitle{Modeling and Verification of a Protocol for
  Operational Support Using Coloured Petri Nets}. In \bibinfo{booktitle}{{\em
  Applications and Theory of Petri Nets - 32nd International Conference,
  {PETRI} {NETS} 2011, Newcastle, UK, June 20-24, 2011. Proceedings}}.
  \bibinfo{pages}{169--188}.
\newblock


\end{thebibliography}

\end{document}